\documentclass[10pt,journal,compsoc]{IEEEtran}

\usepackage{afterpage}
\usepackage{stfloats}
\usepackage{cuted}
\usepackage{caption}
\usepackage{graphicx}
\usepackage{adjustbox}
\usepackage{amsmath}
\usepackage{amssymb}
\usepackage{enumitem}
\usepackage{booktabs}
\usepackage{multirow}
\usepackage[pagebackref=true,breaklinks=true,colorlinks,bookmarks=false,urlcolor=red,citecolor=cyan]{hyperref}
\usepackage{url}
\usepackage{subcaption}
\usepackage{array}
\usepackage{tikz}
\usetikzlibrary{shapes,arrows.meta,positioning,fit,calc}
\usepackage[edges]{forest}
\usepackage{ragged2e}
\usepackage[table]{xcolor}
\definecolor{mygreen}{rgb}{0.0, 0.5, 0.0}
\usepackage{capt-of}

\ifCLASSOPTIONcompsoc
  \usepackage[nocompress]{cite}
\else
  \usepackage{cite}
\fi

\ifCLASSINFOpdf
\else
\fi

\usepackage{wrapfig}

\usepackage[capitalize]{cleveref}

\newcommand{\myparagraph}[1]{\textbf{#1}\hspace{1.8ex}}

\begin{document}
\title{ReMoMask-2: Latent Retrieval-Augmented Masked Motion Generation}

\author{Yiran Wang$^*$,
        Zeyu Zhang$^{*\dag}$,
        Ling Shao, \IEEEmembership{ Fellow,~IEEE,}
        Hao Tang$^\ddag$
\IEEEcompsocitemizethanks{
        \IEEEcompsocthanksitem $^*$Equal contribution. $^\dag$Project lead. \protect
        \IEEEcompsocthanksitem $^\ddag$Corresponding author, E-mail: bjdxtanghao@gmail.com. \protect
        \IEEEcompsocthanksitem Yiran Wang is with the University of Sydney, Camperdown 2006, Australia. \protect
        \IEEEcompsocthanksitem Zeyu Zhang and Hao Tang are with the School of Computer Science, Peking University, Beijing 100871, China. \protect
        \IEEEcompsocthanksitem Ling Shao is with the UCAS-Terminus AI Lab, University of Chinese Academy of Sciences, Beijing 101408, China.\protect
    }%
}

\markboth{Submitted to IEEE Transactions on Pattern Analysis and Machine Intelligence}%
{Wang \MakeLowercase{\textit{et al.}}: ReMoMask-2: Latent Retrieval-Augmented Masked Motion Generation}

\IEEEtitleabstractindextext{%
\justify
\begin{abstract}
Text-to-motion (T2M) generation maps a natural language description to a sequence of human joint movements, offering an intuitive interface for producing human motion in gaming, film production, virtual reality, and robotics. Retrieval-Augmented Text-to-Motion (RAG-T2M) models improve over conventional T2M approaches, particularly on uncommon and complex textual descriptions, by conditioning generation on motion-text pairs retrieved from an external database. However, existing RAG-T2M models remain limited by two challenges. First, retrieval and fusion are structurally inconsistent with motion topology: coarse-grained text-motion retrieval overlooks the hierarchical, part-level structure of human motion, and retrieved evidence is fused by mechanisms that ignore the spatial-temporal structure of the motion latent. Second, retrieved evidence typically resides in a contrastive semantic space learned separately from the latents the generator manipulates, leaving a representation gap between what is retrieved and what is generated. To address the first challenge, we present \textbf{ReMoMask}, a structure-aware RAG framework that couples \textbf{H}ierarchical \textbf{B}idirectional \textbf{M}omentum (HBM) contrastive learning, which employs dual objectives to jointly align global motion semantics and fine-grained part-level features with text; \textbf{S}emantic \textbf{S}patial-\textbf{T}emporal \textbf{A}ttention (SSTA), a topology-aware fusion module that integrates retrieved knowledge via an asymmetric attention mechanism; and \textbf{T}opology \textbf{S}tructured \textbf{M}asking (TSM), a training strategy that adaptively masks motion tokens based on semantic relevance, forcing the model to learn robust part-level grounding. To address the second challenge, we further present \textbf{ReMoMask-2}, which rebuilds the retrieval database directly within the generator's own pre-quantization latent space and aligns text queries to it via a lightweight projector distilled from the retriever, so that the generator consumes the semantic content of the retrieved motion rather than merely registering its presence. Extensive experiments on HumanML3D, KIT-ML, and SnapMoGen demonstrate that our retriever achieves state-of-the-art text-to-motion retrieval, and that ReMoMask-2 attains state-of-the-art generation fidelity among retrieval-augmented approaches and the lowest FID among all compared methods on KIT-ML and SnapMoGen, while its single mask-transformer stage, without any residual-refinement network, surpasses the full two-stage pipeline of ReMoMask and delivers the fastest inference among compared systems.
Code:~\url{https://github.com/AIGeeksGroup/ReMoMask-2}.
Website:~\url{https://aigeeksgroup.github.io/ReMoMask-2}.

\end{abstract}

}

\maketitle

\IEEEdisplaynontitleabstractindextext

\IEEEpeerreviewmaketitle

\IEEEraisesectionheading{\section{Introduction}\label{sec:introduction}}

\IEEEPARstart{H}{uman} motion generation has attracted increasing attention due to its wide applicability in gaming, film production~\cite{motionavatar}, virtual reality, and robotics. By synthesizing realistic and diverse human motions, these methods aim to significantly reduce the cost of manual animation while improving the efficiency and flexibility of content creation. Among various paradigms, text-to-motion (T2M) generation has emerged as a particularly intuitive setting, where a natural language description is directly mapped to a sequence of human joint movements.

\begin{figure}[t!]
    \centering
    \includegraphics[width=\linewidth]{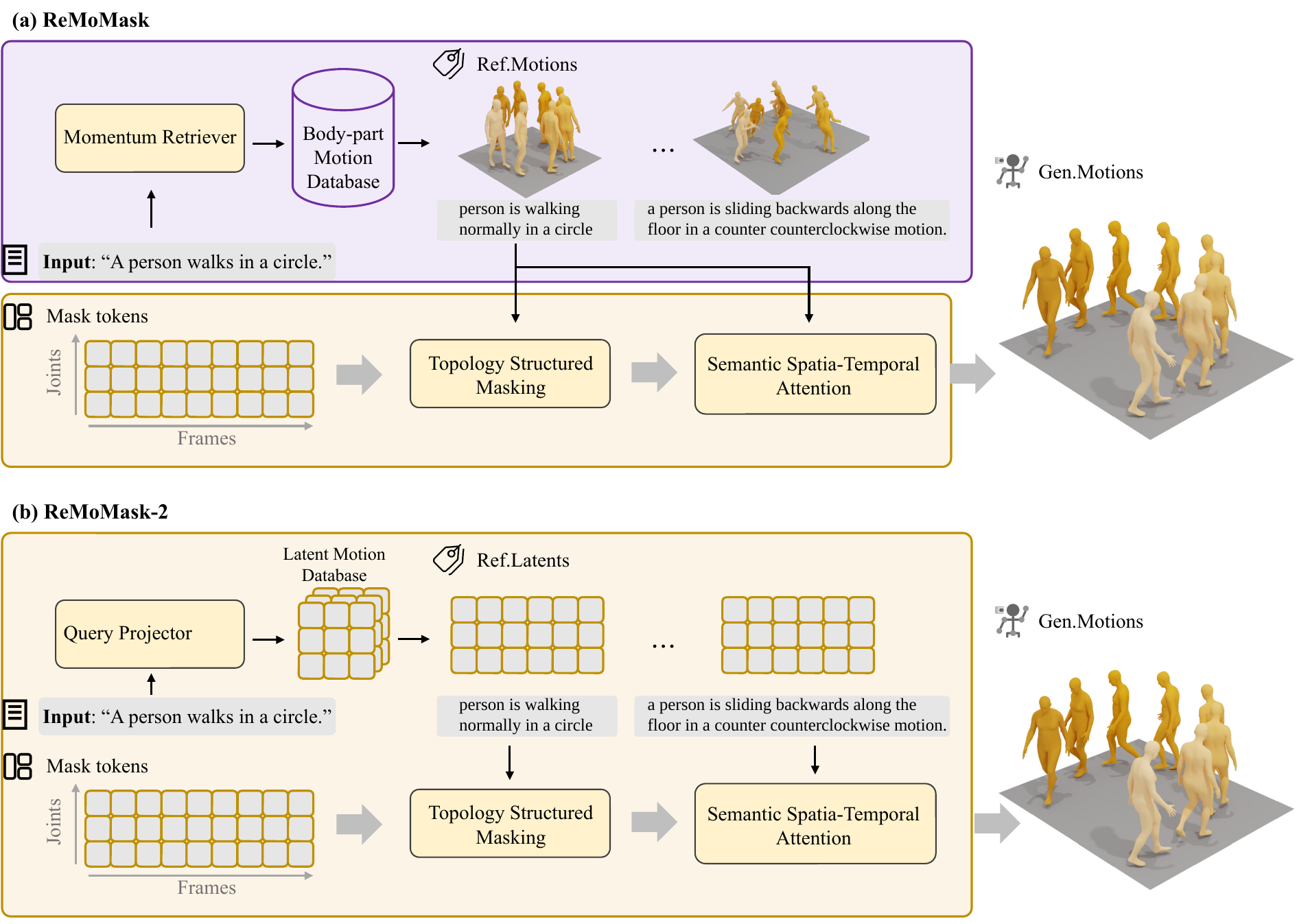}
        \caption{\textbf{ReMoMask vs ReMoMask-2.} (a) ReMoMask retrieves motion clips from a database outside the generator, so every reference crosses the boundary between the two regions. (b) ReMoMask-2 rebuilds the database from the generator's own latents, so store, references and mask tokens share one substrate and nothing crosses.}
        \label{fig:teaser}
\end{figure}

Existing T2M methods can be broadly categorized into two lines of research. The first line consists of \emph{conventional T2M models}, which focus on strengthening the generative backbone itself. Representative approaches include generative adversarial networks (GANs)~\cite{gan,gan1,gan2,gan3,gan4}, variational autoencoders (VAEs)~\cite{vae}, diffusion models~\cite{diffusion}, visual-language models~\cite{cama,vlm1}, motion language models~\cite{motiongpt}, and masked generative models~\cite{mmm,momask,momask2,mogents}. In particular, masked generative models such as MMM~\cite{mmm} and MoMask~\cite{momask} discretize motion sequences into tokenized representations and perform masked token prediction, achieving high-fidelity and temporally coherent motion synthesis.

The second line of work, known as \emph{retrieval-augmented T2M (RAG-T2M)}, enhances generation by retrieving relevant motion–text pairs from an external database and injecting the retrieved evidence into the generative process. By conditioning generation on exemplar motions, RAG-based approaches improve robustness to uncommon or complex textual inputs. Representative methods include ReMoDiffuse~\cite{remodiffuse}, which performs retrieval via text–text similarity using CLIP~\cite{clip}, and ReMoGPT~\cite{remogpt}, which adopts a cross-modal text–motion retriever.

\begin{table}[t]
\centering
\caption{Comparison of different architecture designs.}
\setlength{\tabcolsep}{8pt}
\resizebox{\columnwidth}{!}{
    \begin{tabular}{l | c | c | c | c | c | c }
        \toprule

        \multirow{2}{*}{Method} & \multicolumn{4}{c|}{Retrieval} & \multicolumn{2}{c}{Generation} \\
        \cline{2-7}

        & Global Align & Part Align & Momentum & Space & Fusion & Latent \\
        
        \hline
        \hline

        MDM~\cite{mdm} &  -  & - & - & - & concat & 1D  \\

        T2M-GPT~\cite{t2m-gpt} &  -  & - & - & - & concat & 1D  \\

        MoMask~\cite{momask} &  -  & - & - & - & concat & 1D  \\

        MARDM~\cite{mardm} &  -   & -  & - & - & concat & 1D  \\

        LaMP~\cite{lamp} & -  & - & - & - & cross-attn & 1D  \\

        TMR~\cite{tmr} & $\checkmark$  & $\times$  & $\times$ & Semantic & concat & 1D  \\

        MoRAG~\cite{morag} &  $\times$   & $\times$  & $\times$ & Semantic & cross-attn & 1D  \\

        ReMoGPT~\cite{remogpt} & $\checkmark$  & $\times$  & $\times$ & Semantic & concat & 1D  \\

        ReMoDiffuse~\cite{remodiffuse} & $\checkmark$  & $\times$  & $\times$ & Semantic & cross-attn & 1D  \\

        \midrule
        ReMoMask &  $\checkmark$   & $\checkmark$  & $\checkmark$ & Semantic & SSTA & 2D  \\
        \rowcolor{yellow!20}\textbf{ReMoMask-2} & $\checkmark$ & $\checkmark$ & $\checkmark$ & \textbf{Latent} & SSTA & 2D \\
        \bottomrule
    \end{tabular}
}

\label{tab:delta}
\end{table}

Despite their promising performance, existing RAG-T2M methods implicitly treat retrieval and fusion as independent modules and largely ignore the \emph{structural consistency} between retrieval alignment and motion representation. Through careful analysis, we identify two fundamental design axes of this structural consistency in retrieval-augmented motion generation:

\textbf{(1) Structural Granularity of Alignment.}  
As shown in Table~\ref{tab:delta}, most text–motion retrieval methods rely on contrastive learning to align global motion embeddings with text~\cite{tmr,temos}. However, human motion is inherently hierarchical, organized by skeletal topology and body-part dependencies. Purely global alignment overlooks fine-grained part-level semantics (e.g., left/right limbs or asymmetric actions), limiting retrieval discriminability. Although some works~\cite{remogpt,parco} encode part-level motion features, these representations are typically aggregated without part-level cross-modal alignment, weakening structured correspondence.

\textbf{(2) Structural Compatibility of Fusion.}  
As depicted in Table~\ref{tab:delta}, existing RAG-T2M methods often adopt simple concatenation or vanilla cross-attention, without systematically examining how motion latent structure (e.g., 1D vs. 2D spatial–temporal tokens) interacts with fusion mechanisms. This mismatch between retrieved information and motion representation can limit generation quality.

These observations suggest that performance improvements in RAG-T2M hinge on whether \emph{retrieval alignment and fusion design are structurally consistent with motion topology}, beyond the strength of individual modules. Taken together, these two axes constitute the first challenge we address in this article.

Beyond these two axes, a second challenge remains, along a third design axis orthogonal to the structural ones: in existing RAG-T2M methods~\cite{remodiffuse,remogpt,morag}, the retrieved evidence lives in a contrastive semantic space learned separately from the latents the generator manipulates, so retrieved motions must cross a representation gap before they can guide generation, as reflected by the \emph{Space} column of Table~\ref{tab:delta}. This gap persists even when retrieval alignment and fusion are both made structurally consistent with motion topology.

We propose \textbf{ReMoMask}, a retrieval-augmented masked generative framework for text-to-motion generation. 
To address alignment granularity, we introduce \textbf{H}ierarchical \textbf{B}idirectional \textbf{M}omentum (HBM) alignment, a structured contrastive learning framework that jointly supervises instance-level (global) and part-level bidirectional text–motion correspondence under a momentum-based paradigm. 
To ensure structural compatibility during semantic conditioning, we conduct a systematic study (provided in Section~\ref{sec:preliminary}) on motion latent representations and fusion strategies. 
Our analysis reveals that preserving motion as 2D spatial–temporal tokens, rather than flattening them into 1D sequences, significantly improves semantic injection and generation stability. 
Motivated by this observation, we design \textbf{S}emantic \textbf{S}patial-\textbf{T}emporal \textbf{A}ttention (SSTA), an attention mechanism tailored to inject retrieved motion semantics into the generative backbone. Meanwhile, to further strengthen structural modeling within this 2D formulation, we adopt a \textbf{T}opology \textbf{S}tructured \textbf{M}asking (TSM) strategy, allowing the network to learn richer topological dependencies across body parts and time.

In this paper, we extend our ECCV 2026 conference framework~\cite{remomask}, which inherits this gap, along the third axis of \emph{representation consistency} and present \textbf{ReMoMask-2} (Fig.~\ref{fig:teaser}), which performs retrieval directly in the generator's own pre-quantization latent space $z_e$. Because retrieval keys are then produced by the same frozen encoder that supplies the generative latents, no cross-space translation has to be learned, and retrieved neighbors are directly comparable to the latents the generator predicts.

Our contributions are summarized as follows:

\begin{itemize}
    \item We present \textbf{ReMoMask}, first introduced in our conference version~\cite{remomask}, a retrieval-augmented masked generative framework that enforces structural consistency along the two structural axes identified above: alignment granularity and fusion compatibility. It couples \textbf{HBM}, a hierarchical bidirectional momentum alignment framework that explicitly models both global and part-level text–motion correspondence for fine-grained semantic grounding, with \textbf{SSTA}, a topology-aware spatial–temporal attention mechanism built upon structured 2D motion tokens, and a \textbf{TSM} masking strategy that strengthens topological dependencies across body parts and time.

    \item We present \textbf{ReMoMask-2}, which extends ReMoMask along the third design axis, \emph{representation consistency between the retrieval space and the generative latent space}: it rebuilds the retrieval database in the frozen RVQ-VAE's pre-quantization latent space $z_e$ and aligns text queries to it with a lightweight projector distilled from the HBM retriever, closing the gap between retrieved evidence and the generative substrate.

    \item Extensive experiments on HumanML3D, KIT-ML, and SnapMoGen show that our retriever attains state-of-the-art text-to-motion retrieval on all three benchmarks (R@1 18.49 on HumanML3D, against 11.00 for the next best), and that ReMoMask-2 is the strongest retrieval-augmented generator on all three, with the lowest FID among all compared methods on KIT-ML (0.138) and SnapMoGen (13.174), surpassing the two-stage MoMask pipeline on HumanML3D (FID 0.042 versus 0.046, Top-1 0.528 versus 0.521) with a single generative stage.
\end{itemize}

As shown in Fig.~\ref{fig:teaser}, ReMoMask-2 extends our ReMoMask framework from retrieval in a separate semantic space to retrieval inside the generator's own latent space, which in turn makes a single generative stage sufficient. A preliminary version of this work was accepted to ECCV 2026~\cite{remomask}. That conference version generates motion in two stages, a masked transformer followed by a residual-refinement transformer. The extension consists of two coupled changes: the retrieval database is rebuilt in the generator's pre-quantization latent space $z_e$ and text queries are aligned to it with a lightweight projector distilled from the conference-version retriever, so that retrieved evidence and the generated latents share a single representation; and the residual-refinement transformer is removed, in line with recent single-stage designs~\cite{mmm,parco,momask2}, since the retrieval-conditioned mask transformer already supplies the fine-grained correction that a residual stage is designed to provide. Our experiments bear this out: under a unified 20-repeat protocol on which we reproduce ReMoMask, the single-stage, mask-only ReMoMask-2 surpasses the full mask-plus-residual conference version, achieving a 65.9\% and 75.4\% improvement in FID on HumanML3D ($0.123\!\to\!0.042$) and KIT-ML ($0.562\!\to\!0.138$), respectively, while reducing per-sample inference time from roughly 0.14s to 0.05s, a saving to which the lighter retrieval front end of the latent-aligned design contributes alongside the removed stage; reattaching the residual-refinement transformer not only adds inference cost but actively hurts fidelity, raising FID from 0.042 to 0.068. Thus, ReMoMask-2 treats retrieval as a native part of the generative representation rather than as external evidence that must be translated into it.

\begin{figure*}[t!]
    \centering
    \includegraphics[width=\linewidth]{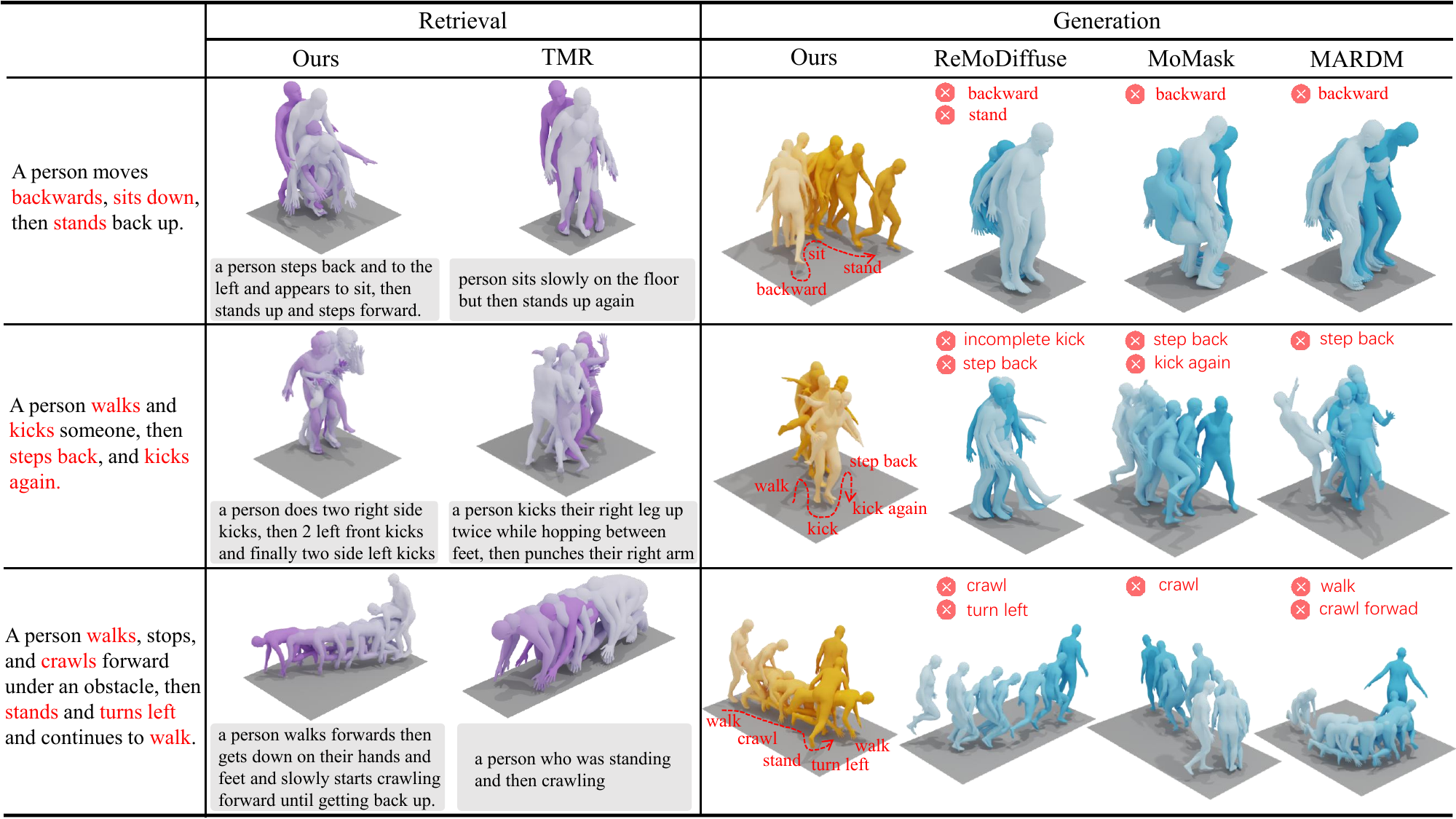}
     \caption{\textbf{Qualitative comparisons.} The darker colors indicate the later in time. The motions generated by our method closely align with the descriptions, outperforming others that exhibit degraded motions or improper semantics.}
        \label{fig:rag_t2m_visual}
\end{figure*}

\section{Related Work}
\label{sec:relatedwork}

\subsection{Text-to-Motion Generation}

Text-to-motion (T2M) generation aims to synthesize realistic human motion sequences conditioned on natural language descriptions. Early approaches explored adversarial learning to establish text–motion correspondence. With the introduction of vector quantization, TM2T~\cite{tm2t} and T2M-GPT~\cite{t2m-gpt} discretized motion sequences and leveraged autoregressive transformers for semantic control. However, autoregressive decoding often suffers from error accumulation.

Recent advances focus on improving motion representation and generation paradigms. MoMask~\cite{momask} introduces hierarchical residual quantization and masked bidirectional transformers for parallel decoding, achieving strong performance on HumanML3D. Diffusion-based methods~\cite{ddpms,motiongpt} further enhance generation quality via non-autoregressive denoising processes. MotionGPT~\cite{motiongpt} unifies multiple generation tasks under a discrete autoregressive framework. 

Beyond backbone improvements, several works investigate fine-grained motion structure modeling. ParCo~\cite{parco} discretizes whole-body motion into part-level components (limbs, backbone, root) to establish structured priors.

\subsection{Retrieval-Augmented Text-to-Motion}

Retrieval-augmented generation (RAG) has been extended beyond NLP to multimodal domains, including motion generation~\cite{rmd,remodiffuse,morag,remogpt}. In retrieval-augmented T2M (RAG-T2M), relevant motion–text pairs are retrieved from an external database and injected into the generative backbone to improve robustness under complex or rare textual conditions. 

Existing approaches typically rely on global contrastive alignment between text and motion embeddings~\cite{tmr,temos}. For instance, ReMoDiffuse~\cite{remodiffuse} performs retrieval based on text–text similarity to indirectly guide generation, while ReMoGPT~\cite{remogpt} introduces part-aware encoders for cross-modal retrieval. However, these methods often aggregate part features without explicit bidirectional cross-modal supervision, leaving fine-grained structural correspondence between text descriptions and specific body parts underexplored. Furthermore, regarding the integration of retrieved information, prior works commonly adopt feature concatenation or standard cross-attention~\cite{remodiffuse,remogpt} without systematically considering the compatibility between the retrieved evidence and the underlying motion latent topology. The impact of motion representation structures on fusion effectiveness remains largely uninvestigated.

\begin{table}[t]
    \centering
    \caption{Comparison of different latent structures and fusion strategies. These pilot results follow the conference evaluation protocol on a MoMask backbone; results under the main protocol used throughout this paper appear in Table~\ref{tab:t2m_experiment}.}
    \label{tab:prelimi_experiment}
    \renewcommand{\arraystretch}{1.5}  %
    \setlength{\tabcolsep}{4pt}  %
    \resizebox{\columnwidth}{!}{
        \begin{tabular}{l| c| c| c| c }
            \toprule
            Generator & Latent & Fusion & FID$\downarrow$ & Top1$\uparrow$ \\
            \hline
            \hline
            
            \multirow{4}{*}{MoMask} & \multirow{2}{*}{1D} & concat    & $0.057^{\pm.013}$ & $0.511^{\pm{.003}}$ \\
            &  & crossAttn & $\underline{0.043^{\pm{.004}}}$ & $\underline{0.525^{\pm{.003}}}$  \\
            \cline{2-5}
             & \multirow{2}{*}{2D} & concat    & $0.049^{\pm{.007}}$ & $0.518^{\pm{.003}}$   \\
            & & crossAttn & $\mathbf{0.036^{\pm .005}}$ & $\mathbf{0.536^{\pm{.002}}}$  \\
            \bottomrule
        \end{tabular}
    }
\end{table}

\section{Design Principles of Retrieval-Augmented Motion Generation}
\label{sec:preliminary}

While retrieval-augmented T2M methods show promise, the interplay between motion representation and fusion strategy remains underexplored. We investigate two key design axes: (1) \emph{motion latent topology} (1D sequence vs. 2D spatial–temporal grid) and (2) \emph{fusion mechanism} (concatenation vs. cross-attention).

Using MoMask~\cite{momask} as a backbone generator, we systematically vary these axes while fixing other components. We construct both 1D ($z_{1d}$) and 2D ($z_{2d}$) latents, integrating retrieved embeddings ($R_m, R_t$ from ReMoDiffuse~\cite{remodiffuse}) via concatenation or cross-attention. All evaluations are performed on the HumanML3D benchmark.

As summarized in Table~\ref{tab:prelimi_experiment}, the optimal configuration combines 2D representation with cross-attention (0.036 FID), the design that directly motivates our \textbf{SSTA} module, which fuses retrieved semantics with motion tokens via structure-aware cross-attention over a 2D latent grid.

\section{Methodology}
\subsection{Overview}
\label{subsec:overview}

\begin{figure*}[t] 
    \centering 
    \includegraphics[width=\textwidth]{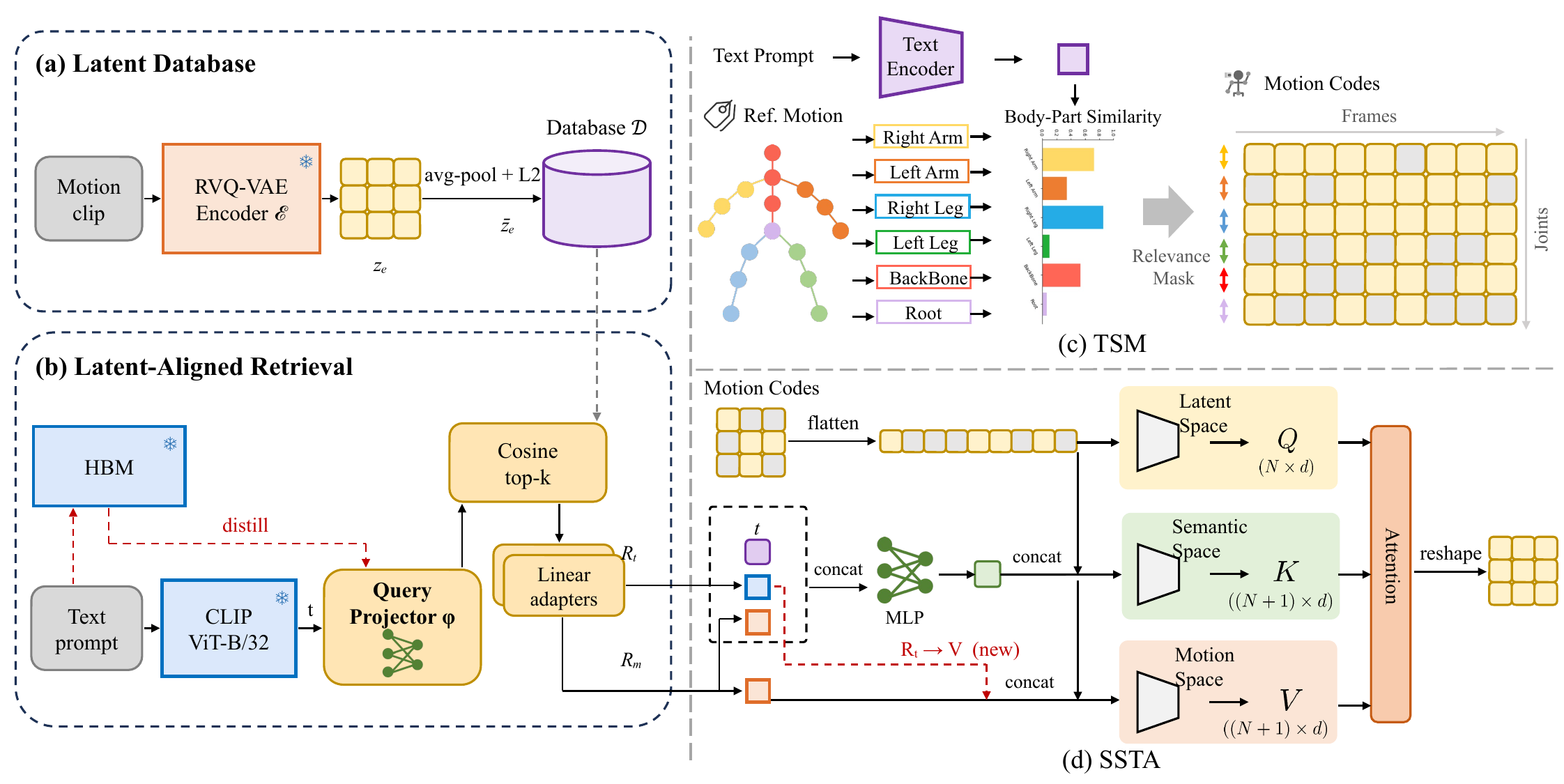} 
    \caption{\textbf{Framework of ReMoMask-2.}
    (a) The retrieval database is built offline by the frozen RVQ-VAE encoder $\mathcal{E}$: each motion becomes a pre-quantisation latent grid, pooled and $\ell_2$-normalised into a key.
    (b) A query projector $\phi$, distilled from the frozen HBM retriever, maps text into that same key space, so cosine retrieval returns evidence already expressed in the generator's representation.
    (c) \textbf{T}opology \textbf{S}tructured \textbf{M}asking (TSM) and (d) \textbf{S}emantic \textbf{S}patial--\textbf{T}emporal \textbf{A}ttention (SSTA) are inherited unchanged from ReMoMask, except that $R_t$ may now enter the Value pathway.
    }
        \label{fig:framework}
\end{figure*}

As illustrated in Fig.~\ref{fig:teaser}, our framework follows a retrieval-augmented generation paradigm. We first employ \textbf{Hierarchical Bidirectional Momentum (HBM)} learning to align text and motion at both instance and part levels, constructing a hierarchically organized embedding space for accurate retrieval. During training, we utilize \textbf{Topology Structured Masking (TSM)}, a structure-aware masked modeling objective that adaptively modulates masking probabilities based on hierarchical relevance to strengthen part-level grounding. Given an input text prompt $x$, relevant motion-text pairs are retrieved from this structured space to obtain reference embeddings ($R_m, R_t$), which are then injected into the generator via our proposed \textbf{Semantic Spatial--Temporal Attention (SSTA)} to enable topology-aware semantic conditioning over 2D motion tokens.

In this pipeline, retrieval operates in the contrastive semantic space constructed by HBM, whereas generation operates on the latents of a 2D RVQ-VAE, two representations learned under disjoint objectives. Section~\ref{subsec:lar} presents the extension that defines \textbf{ReMoMask-2}: we migrate retrieval into the generator's own pre-quantization latent space, so that retrieved evidence arrives, by construction, in the representation the generator natively consumes.

\subsection{Hierarchical Bidirectional Momentum}
\label{subsec:hbm}
\noindent \myparagraph{Hierarchical Motion Decomposition.}
As shown in Fig.~\ref{fig:framework}a, we decompose motion $m$ into $K=6$ semantic parts (e.g., arms, legs, backbone, root), denoted as $m^k$. Given a batch $\{(x_i, m_i)\}_{i=1}^{B}$, we encode text, global motion, and part motions into a shared space:
\begin{equation}
t_i = f_T(x_i), \quad
g_i = f_G(m_i), \quad
p_{i,k} = f_P^{(k)}(m_i^k),
\end{equation}
where $f_T$, $f_G$, and $f_P^{(k)}$ are the respective encoders.

\noindent \myparagraph{Momentum-Based Structured Contrast.}
To stabilize training and enlarge the negative set, we maintain momentum encoders with parameters updated via exponential moving average:
\begin{equation}
\theta^{-} \leftarrow \mu\, \theta^{-} + (1 - \mu)\, \theta,
\end{equation}
where $\theta$ and $\theta^{-}$ denote online and momentum parameters. The resulting momentum embeddings ($t_i^{-}, g_i^{-}, p_{i,k}^{-}$) are stored in a queue $\mathcal{Q}$ serving as negative keys.

\noindent \myparagraph{Contrastive Objective.}
Using cosine similarity $\mathrm{sim}(\cdot,\cdot)$ and temperature $\tau$, we define the positive term $\mathcal{P}(q,k) = \exp(\mathrm{sim}(q,k)/\tau)$ and negative sum $\mathcal{N}(q) = \sum_{k^- \in \mathcal{Q}} \exp(\mathrm{sim}(q,k^-)/\tau)$. The InfoNCE loss is:
\begin{equation}
\mathrm{InfoNCE}(q,k)
=
-\log
\frac{
    \mathcal{P}(q,k)
}{
    \mathcal{P}(q,k) + \mathcal{N}(q)
}.
\end{equation}

\noindent \myparagraph{Bidirectional Alignment.}
We enforce bidirectional alignment at the instance level:
\begin{equation}
\mathcal{L}_{\mathrm{Inst}}
=
\frac{1}{B}
\sum_{i=1}^{B}
\Big(
\mathrm{InfoNCE}(t_i, g_i)
+
\mathrm{InfoNCE}(g_i, t_i)
\Big),
\end{equation}
and further align text with each part representation to preserve fine-grained semantics:
\begin{equation}
\mathcal{L}_{\mathrm{Part}}
=
\frac{1}{BK}
\sum_{i=1}^{B}
\sum_{k=1}^{K}
\Big(
\mathrm{InfoNCE}(t_i, p_{i,k})
+
\mathrm{InfoNCE}(p_{i,k}, t_i)
\Big).
\end{equation}

\noindent \myparagraph{Final Objective.}
The total HBM loss combines global and part-level supervision:
\begin{equation}
\mathcal{L}_{\mathrm{HBM}}
=
\mathcal{L}_{\mathrm{Inst}}
+
\lambda_P \mathcal{L}_{\mathrm{Part}}.
\end{equation}
This hierarchical supervision yields a structurally consistent embedding space for precise retrieval.

\subsection{Part-Level Motion Encoder} 
\label{app:part-level-motion-encoder}
\begin{figure}[t!]
    \centering
     \includegraphics[width=\linewidth]{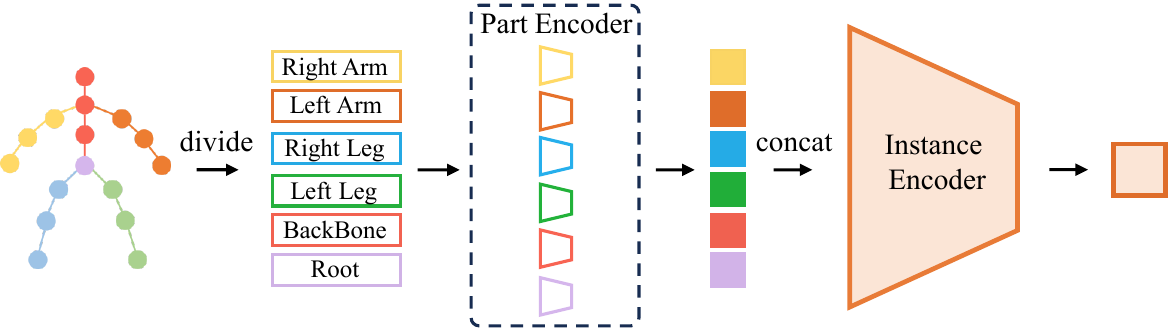}
    \caption{\textbf{Part-level motion encoding for hierarchical alignment.} 
    The full-body motion sequence is first divided into multiple body parts according to the skeletal structure. Each part motion is encoded independently by a shared part encoder to obtain part-level features, which are then concatenated and aggregated by an instance encoder to produce a global motion representation. This hierarchical encoding enables fine-grained part-level semantics while preserving holistic motion information.}
    \label{fig:part-level-motion-encoder}
\end{figure}
To capture the hierarchical structure of human motion, inspired by prior human motion modeling works~\cite{parco,remogpt}, we adopt a part-level motion encoding scheme that decomposes full-body motion into multiple semantically meaningful body parts. As illustrated in Fig.~\ref{fig:part-level-motion-encoder}, a motion sequence is first divided into six parts according to the skeletal topology, including the right arm, left arm, right leg, left leg, backbone, and root. Each part motion is processed independently by a shared part encoder to obtain part-level motion features. These features preserve fine-grained local motion semantics while maintaining parameter efficiency through weight sharing. The resulting part-level representations are then concatenated and aggregated by an instance encoder to form a holistic motion representation, which is subsequently used for hierarchical alignment and retrieval. This hierarchical encoding design enables the model to jointly model local part dynamics and global motion coherence, providing a structured motion representation that facilitates fine-grained text--motion alignment in the proposed framework.

\subsection{Topology Structured Masking}
\label{subsec:tsm}

To fully exploit the hierarchical structure learned by HBM during generation, we introduce \textbf{Topology Structured Masking (TSM)} as a structure-aware training objective. Unlike conventional uniform random masking that ignores semantic relevance, TSM adaptively modulates masking probabilities based on the hierarchical text--motion alignment established by HBM, ensuring the generator focuses on semantically critical body parts.

\noindent \myparagraph{Hierarchical Semantic Relevance.}
As shown in Fig.~\ref{fig:framework}b, given a text embedding $t_i$ and part-level motion embeddings $\{p_{i,k}\}_{k=1}^{K}$ from HBM, we compute part-level relevance weights:
\begin{equation}
\alpha_{i,k}
=
\mathrm{SoftMax}_k
\big(
\mathrm{sim}(t_i, p_{i,k})
\big),
\end{equation}
where $\alpha_{i,k}$ reflects the semantic alignment strength between the input text and the $k$-th body part.

\noindent \myparagraph{Topology-Aware Mask Distribution.}
We define the masking probability for each body part as:
\begin{equation}
\pi_{i,k}
=
\pi_{\mathrm{base}}
\cdot
(1 - \alpha_{i,k}),
\end{equation}
where $\pi_{\mathrm{base}}$ is a global masking ratio. This strategy ensures that semantically aligned parts are preserved as structural anchors, while less relevant parts are masked aggressively. By protecting these core semantic features from corruption, we force the generator to learn how to coordinate global motion conditioned on the key body parts specified by the text, rather than attempting to hallucinate critical semantics from context. These part-level probabilities are then propagated to the 2D spatial--temporal token grid ($J \times T$) according to the skeletal partition, such that all joints belonging to part $k$ share the probability $\pi_{i,k}$.

\noindent \myparagraph{Structured Masked Modeling Objective.}
During training, tokens are masked according to $\pi_{i,k}$ and replaced by a learnable mask token. The generator is optimized to reconstruct the original latent codes $z$:
\begin{equation}
\mathcal{L}_{\mathrm{TSM}}
=
\mathbb{E}_{z}
\left[
\| z_{\mathrm{pred}} - z \|_{\mathrm{masked}}
\right].
\end{equation}
By aligning the masking distribution with semantic relevance, TSM forces the model to focus on reconstructing critical body parts, thereby enhancing part-level grounding and structural consistency.

\subsection{Semantic Spatial-Temporal Attention}
\label{subsec:ssta}
Unlike conventional cross-attention that treats retrieved features as uniform tokens, we design \textbf{Semantic Spatial--Temporal Attention (SSTA)} to explicitly respect the 2D spatial–temporal organization of motion latents through an asymmetric injection mechanism.

We represent motion as a 2D joint–time token grid, flattened into latent sequences $z \in \mathbb{R}^{B \times N \times d}$ (where $N = T \cdot J$). As illustrated in Fig.~\ref{fig:framework}(c), SSTA decouples the roles of semantic conditioning and motion synthesis across the Key and Value pathways.

\noindent \myparagraph{Query: Motion-Centric Focus.}
To preserve the structural integrity of the generation process, the Query is derived solely from the current motion latents:
\begin{equation}
Q = W_q z,
\end{equation}
ensuring that attention computation remains anchored in the spatial–temporal topology of the motion being generated.

\noindent \myparagraph{Semantic Reference Construction.}
We compress the multi-modal retrieval context (prompt $t$, retrieved text $R_t$, and retrieved motion $R_m$) into a compact semantic token:
\begin{equation}
h_{\mathrm{sem}} = \mathrm{MLP}\!\left(
\mathrm{concat}[t;\, R_t;\, R_m]
\right),
\end{equation}
where $h_{\mathrm{sem}} \in \mathbb{R}^{B \times 1 \times d}$ serves as a global modulation signal rather than a sequence of tokens.

\noindent \myparagraph{Key: Topology-Aware Gating.}
The Key pathway integrates both motion states and semantic context to determine \emph{where} to attend:
\begin{equation}
K = W_k \cdot
\mathrm{concat}\!\left(z,\; h_{\mathrm{sem}}\right).
\end{equation}
By injecting $h_{\mathrm{sem}}$ only into the Key, semantic information modulates the attention weights without directly overwriting the motion representation.

\noindent \myparagraph{Value: Motion-Domain Synthesis.}
The Value pathway focuses on \emph{what} information to synthesize, combining current latents with retrieved dynamic priors:
\begin{equation}
V = W_v \cdot
\mathrm{concat}\!\left(z,\; R_m\right).
\end{equation}
Here, $R_m$ provides concrete motion details from the database, while $z$ ensures continuity with the current generation state. Notably, textual semantics ($t, R_t$) are excluded from Value to prevent domain mismatch.

\noindent \myparagraph{Output.}
The final output is computed via standard scaled dot-product attention:
\begin{equation}
\mathrm{Output}
=
\mathrm{SoftMax}
\left(
    \frac{QK^\top}{\sqrt{d}}
\right)V,
\end{equation}
which is then reshaped back to the 2D grid structure. This asymmetric design ensures that retrieved semantics guide the attention focus (via Key) and enrich the motion content (via $R_m$ in Value), while strictly preserving the spatial–temporal topology of the generative backbone.

\subsection{Latent-Aligned Retrieval}
\label{subsec:lar}

The components above enforce structural consistency along the two axes identified in Section~\ref{sec:introduction}: HBM aligns text and motion at the granularity dictated by skeletal topology, and SSTA injects retrieved evidence in a form compatible with the 2D spatial--temporal latent grid. A third axis, however, remains open. In ReMoMask, as in retrieval-augmented T2M at large~\cite{remodiffuse,remogpt,morag}, the space in which evidence is \emph{retrieved} is not the space in which motion is \emph{generated}: HBM embeds text and motion into a contrastive space $\mathcal{S}$, optimized to discriminate matching from non-matching pairs, whereas the masked generator operates on the latent space $\mathcal{Z}$ of the 2D RVQ-VAE, optimized purely for reconstruction. The conference version's retrieval evidence $(R_m, R_t) \in \mathcal{S}$ is therefore consumed by attention layers whose substrate lives in $\mathcal{Z}$: implicitly, the fusion modules are asked to realize a cross-space translation $\psi: \mathcal{S} \rightarrow \mathcal{Z}$, supervised only indirectly through the generation loss. Since the two spaces never interact during training, nothing constrains their geometries to agree: two motions that are close under contrastive semantics may be far apart as latent trajectories. We refer to this discrepancy as the \emph{retrieval--generation representation gap}.

ReMoMask-2 removes the gap at its source: rather than translating evidence across spaces, we construct the retrieval database directly in the generator's own latent space, so that retrieved evidence arrives, by construction, in the representation the generator natively consumes.

\noindent \myparagraph{Pre-Quantization Latent Keys.}
We build retrieval keys with the frozen encoder $\mathcal{E}$ of the pretrained 2D RVQ-VAE~\cite{mogents}. Given a motion $m$, the encoder produces a pre-quantization latent grid
\begin{equation}
z_e = \mathcal{E}(m) \in \mathbb{R}^{d_e \times T' \times J'},
\end{equation}
with channel dimension $d_e = 1024$ over a downsampled grid of $T' = T/4$ temporal steps and $J' = 6$ spatial positions. We deliberately operate on the continuous latent \emph{before} residual quantization rather than on discrete token indices: the pre-quantization latent preserves the continuous geometry of the encoder space and directly supports cosine-based nearest-neighbor search, whereas quantized indices are categorical and discard within-code variation. A motion-level key is obtained by average pooling over the grid followed by $\ell_2$ normalization:
\begin{equation}
\bar{z}_e
=
\frac{\hat{z}_e}{\|\hat{z}_e\|_2},
\qquad
\hat{z}_e
=
\frac{1}{T' J'}
\sum_{t'=1}^{T'}
\sum_{j'=1}^{J'}
z_e(t', j'),
\end{equation}
where $z_e(t', j') \in \mathbb{R}^{d_e}$ denotes the latent vector at grid position $(t', j')$. Applying this to every training motion and expanding each motion over its captions yields the database
\begin{equation}
\mathcal{D}
=
\big\{ \big( \bar{z}_e^{(j)},\, x^{(j)} \big) \big\}_{j=1}^{M},
\qquad
M = 66{,}912,
\end{equation}
covering 23{,}384 training clips (counting mirrored copies as separate database entries); at query time, entries originating from the same motion are deduplicated so that the retrieved set contains distinct exemplars. Since keys are unit-normalized, retrieval reduces to cosine similarity over $\mathcal{D}$. The RVQ-VAE remains entirely frozen throughout: the database is a cached view of representations the generator already uses.

\noindent \myparagraph{Query Projector.}
Retrieval requires mapping a text prompt into the same key space. We instantiate this as a lightweight projector $\phi$ applied to the 512-d CLIP ViT-B/32~\cite{clip} text embedding $t$ already employed by the framework:
\begin{equation}
\phi(t)
=
\mathrm{norm}\big( W_2\, \mathrm{GELU}( W_1 t ) \big),
\end{equation}
where $W_1 \in \mathbb{R}^{d_e \times 512}$, $W_2 \in \mathbb{R}^{d_e \times d_e}$, and $\mathrm{norm}(\cdot)$ denotes $\ell_2$ normalization. The projector holds ${\sim}1.57$M parameters and is the \emph{only} component trained in this stage.

\noindent \myparagraph{Distillation from the HBM Retriever.}
Naively regressing $\phi(t)$ onto the paired key would supervise a single point and ignore the ranking structure that makes retrieval useful. Instead, we distill the HBM retriever of the conference version, which remains a strong cross-modal ranker, into the latent key space, following the soft-target distillation formulation of~\cite{hinton2015distilling}.
For a training caption $x$ with embedding $t$, let $\Omega(x)$ denote the $\kappa$ database entries ranked highest by the teacher ($\kappa = 256$). Teacher and student induce distributions over $\Omega(x)$ from their respective similarities at temperature $\tau = 0.07$, matching the retriever's contrastive temperature:
\begin{equation}
p^{\square}_j
=
\frac{
    \exp\big( s^{\square}_j / \tau \big)
}{
    \sum_{j' \in \Omega(x)} \exp\big( s^{\square}_{j'} / \tau \big)
},
\qquad
\square \in \{ \mathrm{HBM}, \phi \},
\end{equation}
where $s^{\mathrm{HBM}}_j$ is the teacher's text--motion similarity for entry $j$ in $\mathcal{S}$, and $s^{\phi}_j = \mathrm{sim}\big( \phi(t), \bar{z}_e^{(j)} \big)$ is the student's cosine similarity in the latent key space. The projector minimizes the KL divergence from teacher to student,
\begin{equation}
\mathcal{L}_{\mathrm{align}}
=
\mathrm{KL}\big( p^{\mathrm{HBM}} \,\|\, p^{\phi} \big)
=
\sum_{j \in \Omega(x)}
p^{\mathrm{HBM}}_j
\log
\frac{p^{\mathrm{HBM}}_j}{p^{\phi}_j}.
\end{equation}
Restricting the support to the teacher's top-$\kappa$ candidates concentrates supervision on the ranking decisions that matter for retrieval, and the soft targets transfer the teacher's graded preferences rather than a single hard label. The projector is trained for 200 epochs with a batch size of 128, while the RVQ-VAE, the teacher, and the CLIP text encoder all remain frozen. In effect, distillation transfers the teacher's cross-modal ranking into the generative latent space: the resulting retriever is trained to preserve the semantic precision of HBM while returning evidence natively expressed in the generator's representation.

\noindent \myparagraph{Integration with SSTA.}
At generation time, the query $\phi(t)$ retrieves the nearest database entries by cosine similarity, and both retrieval streams are read out of the latent-aligned space: $R_m$ is the pooled key $\bar{z}_e^{(j)}$ of a retrieved entry, and $R_t = \phi(t^{(j)})$ is the projection of the CLIP embedding $t^{(j)}$ of its caption $x^{(j)}$. When $k>1$ entries are retrieved, each stream is averaged over the $k$ entries, so that the fusion pathways always receive a single $R_m$ and a single $R_t$ regardless of $k$. As both streams are $d_e$-dimensional while the generator operates at latent width $d$, we insert two linear adapters, $\mathbb{R}^{d_e} \rightarrow \mathbb{R}^{d}$, one per stream, ahead of the fusion pathways. These adapters are the only architectural addition: SSTA itself---the asymmetric Q-K-V routing of Section~\ref{subsec:ssta}, including the three-way semantic token $h_{\mathrm{sem}} = \mathrm{MLP}(\mathrm{concat}[t;\, R_t;\, R_m])$---is left unchanged. One \emph{routing} decision, however, is revisited as a direct consequence of the alignment. The conference version excluded textual signals from the Value pathway to prevent domain mismatch (Section~\ref{subsec:ssta}): there, $R_t$ was an embedding in $\mathcal{S}$, foreign to the synthesis substrate. After latent alignment, $R_t = \phi(t^{(j)})$ is itself a point in the generator's latent geometry, so the motion-domain criterion governing the Value pathway now admits it; we therefore include $R_t$ alongside $R_m$ in the Value content, while the prompt embedding $t$, still a CLIP-space vector, remains excluded. Latent-aligned retrieval is thus a plug-and-play replacement for the retrieval space; Table~\ref{tab:abl_space}, whose rows share the same generator, fusion modules, and training recipe, isolates the effect of the retrieval representation from that of the fusion design.

\noindent \myparagraph{Single-Stage Deployment.}
ReMoMask-2 is deployed as a single mask-transformer stage. We also equipped it with a residual-refinement stage mirroring the conference pipeline and found the extra stage harmful: FID degrades from 0.042 to 0.068 ($+0.026$), a gap far exceeding the 95\% confidence intervals of either configuration. With retrieval operating in the generative latent space, the coarse token prediction already absorbs the correction that residual refinement is designed to provide, so the residual stage has no systematic error left to remove and its updates act instead as an additional source of perturbation; we therefore omit it.

\section{Experiment}
\label{sec:experiment}
\begin{table*}[t!]
\centering
\caption{\textbf{Results of text-to-motion and motion-to-text retrieval benchmark on HumanML3D, KIT-ML, and SnapMoGen datasets.}
$\dagger$ indicates Results are reproduced by us following the descriptions in the original paper, as no official implementation is publicly available. Best results are highlighted in \textbf{bold}, and second-best results are \underline{underlined}.}
\label{tab:rag_experiment} 
\resizebox{\textwidth}{!}{
    \begin{tabular}{l|l|cccccc|cccccc}
        \toprule
        \multirow{2}{*}{Benchmark} & \multirow{2}{*}{Methods} 
        & \multicolumn{6}{c|}{Text-to-motion retrieval} 
        & \multicolumn{6}{c}{Motion-to-text retrieval} \\
        & 
        & R@1$\uparrow$ & R@2$\uparrow$ & R@3$\uparrow$ & R@5$\uparrow$ & R@10$\uparrow$ & MedR$\downarrow$
        & R@1$\uparrow$ & R@2$\uparrow$ & R@3$\uparrow$ & R@5$\uparrow$ & R@10$\uparrow$ & MedR$\downarrow$ \\
        \hline\hline
        
        \multirow{5}{*}{HumanML3D}
        & TEMOS~\cite{temos}
        & 2.12 & 4.09 & 5.87 & 8.26 & 13.52 & 173.00
        & 3.86 & 4.54 & 6.94 & 9.38 & 14.00 & 183.25 \\
        & TMR~\cite{tmr}
        & 5.68 & 10.59 & 14.04 & 20.34 & 30.94 & 28.00
        & 9.95 & 12.44 & 17.95 & 23.56 & 32.69 & 28.50 \\
        & MotionPatches$^{\dag}$~\cite{motionpatches}
        & 10.80 & 14.98 & 20.00 & 26.72 & 38.02 & 19.00
        & 11.25 & 13.86 & 19.98 & 26.86 & 37.40 & 20.50 \\
        & ReMoGPT$^{\dag}$~\cite{remogpt}
        & \underline{11.00} & \underline{17.02} & \underline{22.18} & \underline{29.48} & \underline{43.43} & \underline{14.00}
        & \underline{12.25} & \underline{14.95} & \underline{21.45} & \underline{28.34} & \underline{39.11} & \underline{19.00} \\
        \rowcolor[gray]{0.90} \cellcolor{white} & HBM (Ours)
        & \textbf{18.49} & \textbf{21.20} & \textbf{25.63} & \textbf{32.40} & \textbf{46.27} & \textbf{12.00}
        & \textbf{14.83} & \textbf{17.63} & \textbf{25.60} & \textbf{30.75} & \textbf{44.61} & \textbf{16.00} \\
        \midrule
        
        \multirow{5}{*}{KIT-ML}
        & TEMOS~\cite{temos}
        & 7.11 & 13.25 & 17.59 & 24.10 & 35.66 & 24.00
        & 11.69 & 15.30 & 20.12 & 26.63 & 36.39 & 26.50 \\
        & TMR~\cite{tmr}
        & 7.23 & 13.98 & 20.36 & 28.31 & 40.12 & 17.00
        & 11.20 & 13.86 & 20.12 & 28.07 & 38.55 & 18.00 \\
        & MotionPatches$^{\dag}$~\cite{motionpatches}
        & \underline{14.02} & \underline{21.08} & \underline{28.91} & \underline{34.10} & \underline{50.00} & \underline{10.50}
        & \textbf{13.61} & \textbf{17.26} & \textbf{27.54} & \underline{33.33} & \underline{44.77} & \textbf{13.00} \\
        & ReMoGPT$^{\dag}$~\cite{remogpt}
        & 13.42 & 17.65 & 22.24 & 32.30 & 44.68 & 12.50
        & 11.93 & 14.12 & 21.59 & 30.54 & 39.68 & 16.50 \\
        \rowcolor[gray]{0.90} \cellcolor{white} & HBM (Ours)
        & \textbf{16.75} & \textbf{23.10} & \textbf{30.50} & \textbf{37.56} & \textbf{52.20} & \textbf{8.50}
        & \underline{12.14} & \underline{16.36} & \underline{25.64} & \textbf{35.58} & \textbf{46.15} & \underline{14.00} \\
        \midrule
        
        \multirow{5}{*}{SnapMoGen}
        & TEMOS~\cite{temos} & 5.03 & 8.51 & 10.79 & 14.48 & 20.36 & 42.00
        & 4.72 & 6.67 & 7.95 & 10.11 & 16.74 & 64.00 \\
        & TMR~\cite{tmr} & 7.29 & 11.80 & 14.43 & 18.82 & 27.39 & 28.50
        & 8.03 & 10.22 & 13.49 & 20.91 & 29.06 & 33.00 \\
        & MotionPatches$^{\dag}$~\cite{motionpatches} & \underline{8.92} & \underline{14.75} & \underline{18.13} & \underline{25.46} & \underline{34.95} & \underline{19.00}
        & 9.08 & 12.24 & 14.73 & 22.42 & 33.52 & 25.50 \\
        & ReMoGPT$^{\dag}$~\cite{remogpt} & 8.15 & 13.27 & 16.05 & 24.72 & 33.48 & 20.50 
        & \underline{9.94} & \underline{13.55} & \underline{16.41} & \underline{23.87} & \underline{35.53} & \underline{23.00} \\
        \rowcolor[gray]{0.90} \cellcolor{white} & HBM (Ours) & \textbf{12.43} & \textbf{17.84} & \textbf{22.54} & \textbf{28.22} & \textbf{42.83} & \textbf{16.50}
        & \textbf{11.62} & \textbf{15.35} & \textbf{19.32} & \textbf{26.18} & \textbf{39.41} & \textbf{19.00} \\
        \bottomrule
    \end{tabular}
}
\end{table*}

\begin{table*}[t]
    \centering
    \caption{\textbf{Quantitative evaluation on HumanML3D, KIT-ML, and SnapMoGen datasets.} We repeat the evaluation 20 times and report the average with 95\% confidence interval. Results marked with $\dagger$ are reproduced following the original paper, as no official implementation is publicly available. Within each framework category, \textbf{bold} and \underline{underline} indicate the best and the second best results.}
    \label{tab:t2m_experiment} 
    \resizebox{\textwidth}{!}{
    \begin{tabular}{l|l| c | c c c| c |c| c| c}
        \toprule
          & \multirow{2}{*}{Methods} & \multirow{2}{*}{Framework}  & \multicolumn{3}{c|}{R-Precision$\uparrow$} & \multirow{2}{*}{FID$\downarrow$} & \multirow{2}{*}{MMDIST$\downarrow$} & \multirow{2}{*}{Diversity$\rightarrow$} & \multirow{2}{*}{MModality$\uparrow$}\\
        \cline{4-6}
           ~& ~ & ~& Top 1 & Top 2 & Top 3 & & & & \\

        \hline\hline 
        \rule{0pt}{2.5ex} %
        \multirow{14}{*}{\rotatebox{90}{\textbf{HumanML3D}}} &
        Real & - & $0.511^{\pm.003}$ & $0.703^{\pm.003}$  & $0.797^{\pm.002}$ &$0.002^{\pm0.000}$ & $2.974^{\pm0.008}$& $9.503^{\pm0.065}$ & $-$ \\
        \cmidrule{2-10}
        & MDM~\cite{mdm} & \multirow{7}{*}{t2m} & $-$ & $-$  & $0.611^{\pm.007}$ &$0.544^{\pm.044}$ & $5.566^{\pm.027}$& $\mathbf{9.559^{\pm.086}}$ & $\mathbf{2.799^{\pm.072}}$ \\
        & T2M-GPT~\cite{t2m-gpt} & &$0.492^{\pm.003}$&$0.679^{\pm.002}$&$0.775^{\pm.002}$&$0.141^{\pm.005}$&$3.121^{\pm.009}$& $9.761^{\pm.073}$ & $1.856^{\pm.003}$ \\ 
        & MoMask~\cite{momask} & &$0.521^{\pm.003}$ & $0.710^{\pm.002}$  &$0.805^{\pm.001}$ & $0.046^{\pm.002}$ &$2.969^{\pm.009}$& $9.628^{\pm.070}$ & $1.245^{\pm.041}$\\
        & MoGenTS~\cite{mogents} & &$\underline{0.529^{\pm.003}}$ & $\underline{0.719^{\pm.002}}$& $\underline{0.812^{\pm.002}}$& $\underline{0.033^{\pm.001}}$ & $\underline{2.867^{\pm.006}}$& $\underline{9.570^{\pm.077}}$ & $1.205^{\pm.043}$\\
        & MoMask++~\cite{momask2} & &$0.517^{\pm.002}$  &$0.709^{\pm.002}$ &$0.803^{\pm.002}$ &$0.069^{\pm.003}$&$2.948^{\pm.007}$ & $9.611^{\pm.032}$ & $1.192^{\pm.053}$\\
        & MARDM~\cite{mardm} & &$0.500^{\pm.004}$ &$0.695^{\pm.003}$&$0.795^{\pm.003}$&$0.114^{\pm.007}$&$2.945^{\pm.004}$&$9.718^{\pm.031}$ & $2.231^{\pm.071}$\\
        & LaMP~\cite{lamp} & &$\mathbf{0.557^{\pm.003}}$ &$\mathbf{0.751^{\pm.002}}$&$\mathbf{0.843^{\pm.001}}$& $\mathbf{0.032^{\pm.002}}$ &$\mathbf{2.759^{\pm.007}}$&$9.571^{\pm.069}$ & $\underline{2.794^{\pm.041}}$\\
        \cline{2-10}
        \rule{0pt}{2.5ex} %
        & ReMoDiffuse~\cite{remodiffuse} & \multirow{6}{*}{RAG-t2m} &$0.510^{\pm.005}$& $0.698^{\pm.006}$& $0.795^{\pm.004}$ & $\underline{0.103^{\pm.004}}$&$2.974^{\pm.016}$& $9.018^{\pm.075}$&$1.795^{\pm.043}$\\
        & ReMoGPT$^{\dag}$~\cite{remogpt} & &$0.501^{\pm.003}$& $0.688^{\pm.003}$& $0.792^{\pm.005}$ & $0.205^{\pm.023}$&$2.929^{\pm.020}$& $9.763^{\pm.056}$&$\mathbf{2.816^{\pm.003}}$\\
        & RMD$^{\dag}$~\cite{rmd} & &$\underline{0.524^{\pm.002}}$& $\underline{0.715^{\pm.002}}$& $\underline{0.811^{\pm.001}}$ & $0.111^{\pm.005}$&$\underline{2.879^{\pm.006}}$& $\mathbf{9.527^{\pm.090}}$&$2.604^{\pm.084}$\\
        & MoRAG-Diffuse$^{\dag}$~\cite{morag} & &$0.511^{\pm.003}$& $0.699^{\pm.003}$& $0.792^{\pm.002}$ & $0.270^{\pm.010}$&$2.950^{\pm.001}$& $9.536^{\pm.104}$&$\underline{2.773^{\pm.011}}$\\
        & ReMoMask~\cite{remomask} && $0.485^{\pm.003}$ & $0.676^{\pm.003}$ & $0.777^{\pm.002}$ & $0.123^{\pm.003}$ & $3.090^{\pm.008}$ & $9.357^{\pm.081}$ & $1.446^{\pm.045}$ \\
        \rowcolor{yellow!20} \cellcolor{white} & \textbf{ReMoMask-2 (Ours)} & & $\mathbf{0.528^{\pm.002}}$ & $\mathbf{0.719^{\pm.002}}$ & $\mathbf{0.813^{\pm.002}}$ & $\mathbf{0.042^{\pm.002}}$ & $\mathbf{2.860^{\pm.009}}$ & $\underline{9.472^{\pm.079}}$ & $1.243^{\pm.038}$ \\

        \midrule
        \rule{0pt}{1.5ex} %
        \multirow{14}{*}{\rotatebox{90}{\textbf{KIT-ML}}} &
        Real & - & $0.424^{\pm.005}$ & $0.649^{\pm.006}$  & $0.779^{\pm.006}$ &$0.031^{\pm0.004}$ & $2.788^{\pm0.012}$& $11.080^{\pm.097}$ & $-$ \\
        \cmidrule{2-10}
        & MDM~\cite{mdm} & \multirow{7}{*}{t2m} & $-$ & $-$  & $0.396^{\pm.004}$ &$0.497^{\pm.021}$ & $9.191^{\pm.022}$& $10.847^{\pm.109}$ & $\underline{1.907^{\pm.214}}$ \\
        & T2M-GPT~\cite{t2m-gpt} & &$0.416^{\pm.006}$&$0.627^{\pm.006}$&$0.745^{\pm.006}$&$0.514^{\pm.029}$&$3.007^{\pm.023}$& $10.860^{\pm.049}$ & $1.798^{\pm.157}$ \\ 
        & MoMask~\cite{momask} & &$0.433^{\pm.007}$ & $0.656^{\pm.005}$  &$0.781^{\pm.005}$ & $0.204^{\pm.011}$ &$2.779^{\pm.022}$& $10.203^{\pm.038}$ & $1.131^{\pm.043}$\\
        & MoGenTS~\cite{mogents} & &$\underline{0.445^{\pm.006}}$ & $\underline{0.671^{\pm.006}}$& $\underline{0.797^{\pm.005}}$& $\underline{0.143^{\pm.004}}$ & $\underline{2.711^{\pm.024}}$& $10.918^{\pm.090}$ & $1.493^{\pm.024}$\\
        & MoMask++~\cite{momask2} & &$0.428^{\pm.008}$ &$0.628^{\pm.006}$&$0.755^{\pm.007}$&$0.194^{\pm.016}$&$2.814^{\pm.006}$&$\underline{10.927^{\pm.101}}$ & $1.247^{\pm.040}$\\
        & MARDM~\cite{mardm} & &$0.387^{\pm.006}$ &$0.610^{\pm.006}$&$0.749^{\pm.006}$&$0.242^{\pm.014}$&$2.970^{\pm.032}$&$10.891^{\pm.045}$ & $1.312^{\pm.053}$\\
        & LaMP~\cite{lamp} & &$\mathbf{0.479^{\pm.006}}$ &$\mathbf{0.691^{\pm.005}}$&$\mathbf{0.826^{\pm.005}}$ & $\mathbf{0.141^{\pm.013}}$ & $\mathbf{2.704^{\pm.018}}$ & $\mathbf{10.929^{\pm.101}}$ & $\mathbf{1.973^{\pm.071}}$\\
        \cline{2-10}
        \rule{0pt}{2.5ex} %
        & ReMoDiffuse~\cite{remodiffuse} & \multirow{6}{*}{RAG-t2m} &$0.427^{\pm.014}$& $0.641^{\pm.004}$& $0.765^{\pm.055}$ & $\underline{0.155^{\pm.006}}$&$\underline{2.814^{\pm.012}}$& $10.800^{\pm.105}$&$1.239^{\pm.028}$\\
        & ReMoGPT$^{\dag}$~\cite{remogpt} & &$0.376^{\pm.006}$& $0.585^{\pm.004}$& $0.734^{\pm.005}$ & $0.425^{\pm.022}$&$2.935^{\pm.009}$& $10.712^{\pm.092}$&$\underline{1.362^{\pm.020}}$\\
        & RMD$^{\dag}$~\cite{rmd} & &$\underline{0.433^{\pm.007}}$& $\underline{0.652^{\pm.005}}$& $\underline{0.776^{\pm.004}}$ & $0.320^{\pm.019}$&$2.863^{\pm.015}$& $\underline{10.879^{\pm.110}}$&$\mathbf{1.364^{\pm.100}}$\\
        & MoRAG-Diffuse$^{\dag}$~\cite{morag} & &$0.330^{\pm.008}$& $0.542^{\pm.007}$& $0.675^{\pm.004}$ & $0.614^{\pm.043}$&$3.251^{\pm.022}$& $10.244^{\pm.092}$&$1.268^{\pm.045}$\\
        & ReMoMask~\cite{remomask} && $0.404^{\pm.007}$ & $0.619^{\pm.005}$ & $0.745^{\pm.005}$ & $0.562^{\pm.019}$ & $3.027^{\pm.018}$ & $10.508^{\pm.105}$ & $0.792^{\pm.030}$ \\
        \rowcolor{yellow!20} \cellcolor{white} & \textbf{ReMoMask-2 (Ours)} & & $\mathbf{0.457^{\pm.006}}$ & $\mathbf{0.673^{\pm.005}}$ & $\mathbf{0.801^{\pm.004}}$ & $\mathbf{0.138^{\pm.007}}$ & $\mathbf{2.743^{\pm.016}}$ & $\mathbf{10.964^{\pm.108}}$ & $0.947^{\pm.043}$ \\

        \midrule
        \rule{0pt}{1.5ex} %
        \multirow{14}{*}{\rotatebox{90}{\textbf{SnapMoGen}}} &
        Real & - & $0.940^{\pm.001}$ & $0.976^{\pm.001}$  & $0.985^{\pm.001}$ &$0.001^{\pm.000}$ & $2.835^{\pm.001}$& $19.849^{\pm.063}$ & $-$ \\
        \cmidrule{2-10}
        & MDM~\cite{mdm} & \multirow{7}{*}{t2m} &$0.503^{\pm.002}$& $0.653^{\pm.002}$& $0.727^{\pm.002}$ & $57.783^{\pm.092}$ & $8.627^{\pm.012}$ & $18.701^{\pm.043}$ & $\mathbf{13.412^{\pm.231}}$\\
        & T2M-GPT~\cite{t2m-gpt} & &$0.618^{\pm.002}$& $0.773^{\pm.002}$& $0.812^{\pm.002}$ & $32.629^{\pm.087}$&$6.855^{\pm.032}$& $18.513^{\pm.035}$ & $9.172^{\pm.181}$\\
        & MoMask~\cite{momask} & &$\underline{0.777^{\pm.002}}$& $\underline{0.888^{\pm.002}}$& $\underline{0.927^{\pm.002}}$ & $\underline{17.404^{\pm.051}}$&$3.812^{\pm{.008}}$& $19.583^{\pm.074}$ & $8.183^{\pm.184}$\\
        & MoGenTS~\cite{mogents} & &$0.742^{\pm.004}$& $0.875^{\pm.003}$& $0.912^{\pm.004}$ & $23.927^{\pm.045}$ & $3.923^{\pm.006}$& $19.457^{\pm.061}$ &$8.841^{\pm{.197}}$\\
        & MoMask++~\cite{momask2} & &$\mathbf{0.802^{\pm.001}}$& $\mathbf{0.905^{\pm.002}}$& $\mathbf{0.938^{\pm.001}}$ & $\mathbf{15.060^{\pm.065}}$&$\underline{3.784^{\pm{.006}}}$& $\mathbf{19.764^{\pm{.027}}}$&$7.259^{\pm.180}$\\
        & MARDM~\cite{mardm} & &$0.659^{\pm.002}$& $0.812^{\pm.002}$& $0.860^{\pm.002}$ & $26.878^{\pm.131}$&$3.947^{\pm.009}$& $19.598^{\pm.019}$ & $\underline{9.812^{\pm.287}}$\\
        & LaMP~\cite{lamp} & &$0.711^{\pm.005}$ &$0.803^{\pm.006}$&$0.900^{\pm.003}$&$19.114^{\pm.085}$ & $\mathbf{3.697^{\pm.014}}$ & $\underline{19.618^{\pm.037}}$ & $9.325^{\pm.239}$\\
        \cline{2-10}
        \rule{0pt}{2.5ex} %
        & ReMoDiffuse~\cite{remodiffuse} & \multirow{6}{*}{RAG-t2m}  &$0.491^{\pm.003}$& $0.673^{\pm.007}$& $0.785^{\pm.002}$ & $68.460^{\pm.091}$ & $5.930^{\pm{.029}}$ & $18.753^{\pm.046}$ & $8.260^{\pm{.215}}$\\
        & ReMoGPT$^{\dag}$~\cite{remogpt} & &$0.647^{\pm.002}$& $0.793^{\pm.004}$& $0.839^{\pm.002}$ & $44.121^{\pm.016}$&$4.883^{\pm{.008}}$& $19.435^{\pm.051}$ & $\mathbf{9.347^{\pm{.459}}}$\\
        & RMD$^{\dag}$~\cite{rmd} & &$0.534^{\pm.002}$& $0.658^{\pm.002}$& $0.713^{\pm.003}$ & $49.084^{\pm.023}$ & $5.916^{\pm{.017}}$& $18.566^{\pm.021}$ & $7.559^{\pm{.273}}$\\
        & MoRAG-Diffuse$^{\dag}$~\cite{morag} & &$0.597^{\pm.003}$& $0.792^{\pm.002}$& $0.835^{\pm.002}$ & $41.417^{\pm.030}$ & $5.924^{\pm.036}$ & $18.927^{\pm.032}$ & $8.278^{\pm{.247}}$\\
        & ReMoMask~\cite{remomask} && $\underline{0.752^{\pm.003}}$ & $\underline{0.883^{\pm.002}}$ & $\underline{0.921^{\pm.002}}$ & $\underline{27.816^{\pm.226}}$ & $\underline{3.712^{\pm.011}}$ & $\underline{19.472^{\pm.058}}$ & $\underline{8.613^{\pm.192}}$ \\
        \rowcolor{yellow!20} \cellcolor{white} & \textbf{ReMoMask-2 (Ours)} & & $\mathbf{0.788^{\pm.003}}$ & $\mathbf{0.897^{\pm.004}}$ & $\mathbf{0.934^{\pm.002}}$ & $\mathbf{13.174^{\pm.057}}$ & $\mathbf{3.706^{\pm.011}}$ & $\mathbf{19.703^{\pm.046}}$ & $8.331^{\pm.174}$ \\
        \bottomrule
    \end{tabular}}
    \vspace{-1em}
\end{table*}

\begin{figure*}[t!]
    \centering
    \includegraphics[width=\linewidth]{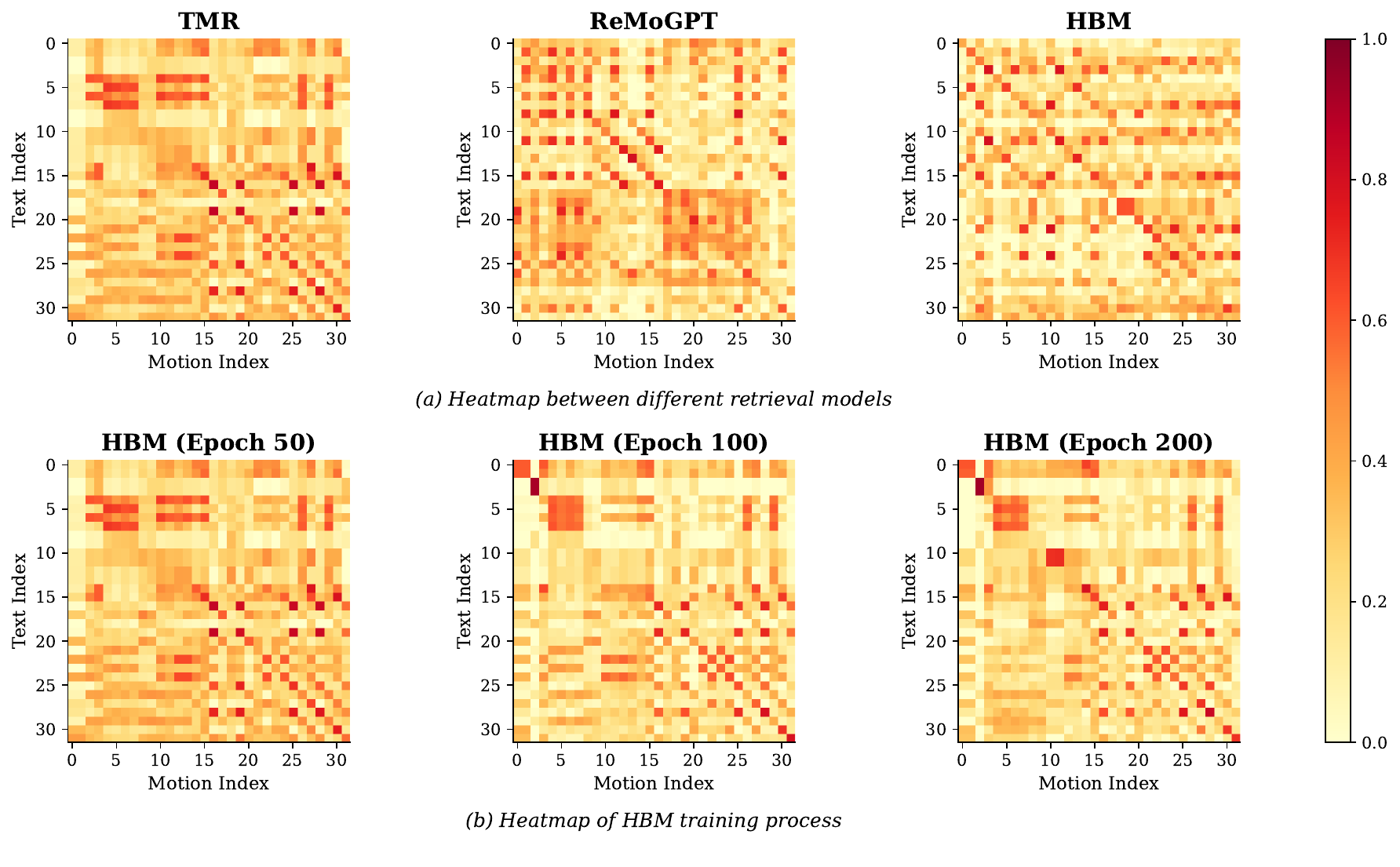}
        \caption{\textbf{Heatmap of similarity matrix.} The diagonal represents positive pairs, with darker colors meaning higher semantic similarity, conducted on HumanML3D.}
        \label{fig:heatmap}
\end{figure*}

\subsection{Dataset and Evaluation Metrics}
We evaluate our model on HumanML3D~\cite{humanml3d}, KIT-ML~\cite{kitml}, and SnapMoGen~\cite{momask2} datasets. The HumanML3D consists of 14,616 motions and 44,970 texts, while KIT-ML consists of 3,911 motions and 6,278 texts. The SnapMoGen is the largest available dataset, comprising 20,450 motions and 122,565 texts. The motion feature dimensions are 263, 251, and 296 for HumanML3D, KIT-ML, and SnapMoGen, respectively. HumanML3D and KIT-ML are split into training, validation, and test sets with a ratio of \textbf{0.80:0.05:0.15}, while SnapMoGen uses a split of \textbf{0.85:0.05:0.10}.

\noindent \myparagraph{Evaluation Metrics.}
For text–motion retrieval, we report Recall at different ranks (R@1, R@2, R@3, R@5, and R@10), which measures the proportion of queries whose ground-truth match appears within the top-$k$ retrieved results. We also report the median rank (MedR), where lower values indicate better retrieval performance.
For motion generation, following prior work~\cite{momask,mogents}, we evaluate text-to-motion (T2M) performance from three perspectives: (1) motion quality, measured by the Frechet Inception Distance (FID); (2) generation diversity, evaluated using Diversity and Multi-Modality (MModality); and (3) text–motion alignment, assessed by R-Precision at top 1/2/3 and the Multi-Modal Distance (MM Dist). 
Formal definitions of each metric are provided below.

We report standard evaluation metrics widely adopted in text-to-motion generation following prior work~\cite{humanml3d}. All metrics are computed in a shared embedding space obtained from the pretrained evaluation networks introduced in~\cite{humanml3d}. We denote the feature representations of ground-truth motions, generated motions, and text descriptions as $f_{gt}$, $f_{pred}$, and $f_{text}$, respectively.

\paragraph{Frechet Inception Distance (FID).}
FID evaluates the distribution-level similarity between generated motions and ground-truth motions in the feature space. It is defined as

\begin{equation}
\begin{aligned}
\text{FID} = {} & \lVert \mu_{gt} - \mu_{pred} \rVert^2 \\
& + \mathrm{Tr}\!\left(
\Sigma_{gt} + \Sigma_{pred}
- 2(\Sigma_{gt}\Sigma_{pred})^{\frac{1}{2}}
\right)
\end{aligned}
\label{formula:fid}
\end{equation}

where $\mu_{gt}$ and $\mu_{pred}$ denote the empirical means of $f_{gt}$ and $f_{pred}$, and $\Sigma_{gt}$ and $\Sigma_{pred}$ are the corresponding covariance matrices.

\paragraph{R-Precision (Top-$k$).}
R-Precision assesses text--motion matching accuracy. For each generated motion, its ground-truth text description is combined with 31 randomly sampled mismatched descriptions from the test set to form a candidate pool. We rank all candidates according to the Euclidean distance between motion and text features, and report the retrieval accuracy at Top-1, Top-2, and Top-3. A retrieval is considered successful if the ground-truth description appears within the top-$k$ ranked results.

\paragraph{Multimodal Distance (MM-Dist).}
MM-Dist measures the alignment between generated motions and their corresponding text descriptions at the feature level. Given $N$ text--motion pairs, it is computed as the average Euclidean distance between motion and text features:
\begin{equation}
\text{MM-Dist} = \frac{1}{N}\sum_{i=1}^{N}\lVert f_{pred,i} - f_{text,i} \rVert .
\label{formula:mm-dis}
\end{equation}

\paragraph{Diversity.}
Diversity quantifies the overall variability of generated motions across the dataset. We randomly sample $S_{dis}$ pairs of generated motion features, denoted as $(f_{pred,i}, f'_{pred,i})$, and compute
\begin{equation}
\text{Diversity} = \frac{1}{S_{dis}}\sum_{i=1}^{S_{dis}} \lVert f_{pred,i} - f'_{pred,i} \rVert .
\label{formula:diversity}
\end{equation}
Following~\cite{humanml3d}, we set $S_{dis}=300$ in all experiments.

\paragraph{Multimodality (MModality).}
MModality evaluates the diversity of motions generated from the same text description. For each text input, we generate multiple motion samples and randomly select two subsets, each containing 10 motions. Let $(f_{pred,i,j}, f'_{pred,i,j})$ denote the feature pair from the $j$-th sample of the $i$-th text. MModality is computed as
\begin{equation}
\text{MModality} = \frac{1}{10N}\sum_{i=1}^{N}\sum_{j=1}^{10}\lVert f_{pred,i,j} - f'_{pred,i,j} \rVert .
\label{formula:mmodality}
\end{equation}

\subsection{Implementation Details}
\label{subsec:impl}
For the motion representation, we use a pretrained 2D-RVQ-VAE~\cite{mogents} comprising two 6-layer residual-quantization branches: a joint-level 2D branch with codebooks of 256 codes of 1024 dimensions, whose pre-quantization latent provides our retrieval keys $z_e$, and a holistic 1D branch with codebooks of 512 codes of 512 dimensions.
For the topology structured masking, we set $\pi_{base}$ to 0.5.
For the text–motion retrieval model, the motion encoder employs a 4-layer transformer for each body part, with both part-level and instance-level motion embeddings set to 512 dimensions. 
The text encoder is a frozen pretrained CLIP~\cite{clip} ViT-B/32 model with one additional trainable transformer layer, shared across all models. 
For the motion masked model, the SSTA is set to have 6 layers, 8 heads, and 512 latent dimensions. The retrieval model is trained for 200 epochs on a Tesla A800 GPU, with a $\lambda_{P}$ of 1, a batch size of 128, a queue size of 65,536, and a momentum $\mu$ of 0.999.
The masked models are trained on 8 Tesla A800 GPUs for up to 2,000 epochs with a batch size of 64. All models are implemented in PyTorch.

\noindent \myparagraph{Evaluation Protocol.}
All results in Table~\ref{tab:t2m_experiment} are obtained under a unified protocol: metrics are computed by the same evaluation pipeline, repeated 20 times, and reported with 95\% confidence intervals; reconstruction measurements, which are deterministic, are the only exception and are reported without intervals. Entries marked $\dagger$ are reproduced by us following the original description, as no public implementation is available. For our conference-version ReMoMask, the table reports our own reproduction under this protocol, obtained by re-running the publicly released checkpoints where available and retraining from the official configuration otherwise. ReMoMask-2 is evaluated under the same protocol throughout.

\subsection{Main Results}
\noindent \myparagraph{Evaluation of Motion Retrieval.}
We evaluate ReMoMask on text-motion and motion-text retrieval on HumanML3D, KIT-ML, and SnapMoGen (Table~\ref{tab:rag_experiment}), where it attains state-of-the-art text-to-motion retrieval on all three benchmarks and the best motion-to-text retrieval on HumanML3D and SnapMoGen.
ReMoMask consistently outperforms prior methods on HumanML3D and SnapMoGen, improving text-to-motion R@1 on HumanML3D to 18.49 versus 11.00 for the next-best method, and increasing R@1 on SnapMoGen from 8.92 to 12.43. On KIT-ML, ReMoMask is best on all six text-to-motion metrics, while in the motion-to-text direction it leads on R@5 and R@10 and is second on the remaining four metrics.
Fig.~\ref{fig:heatmap} further shows that HBM yields clearer diagonal structures with suppressed off-diagonal responses, visualizing the same discriminative text–motion alignment reflected in the retrieval metrics above.

\begin{figure}[t]
    \centering
    \includegraphics[width=0.80\linewidth]{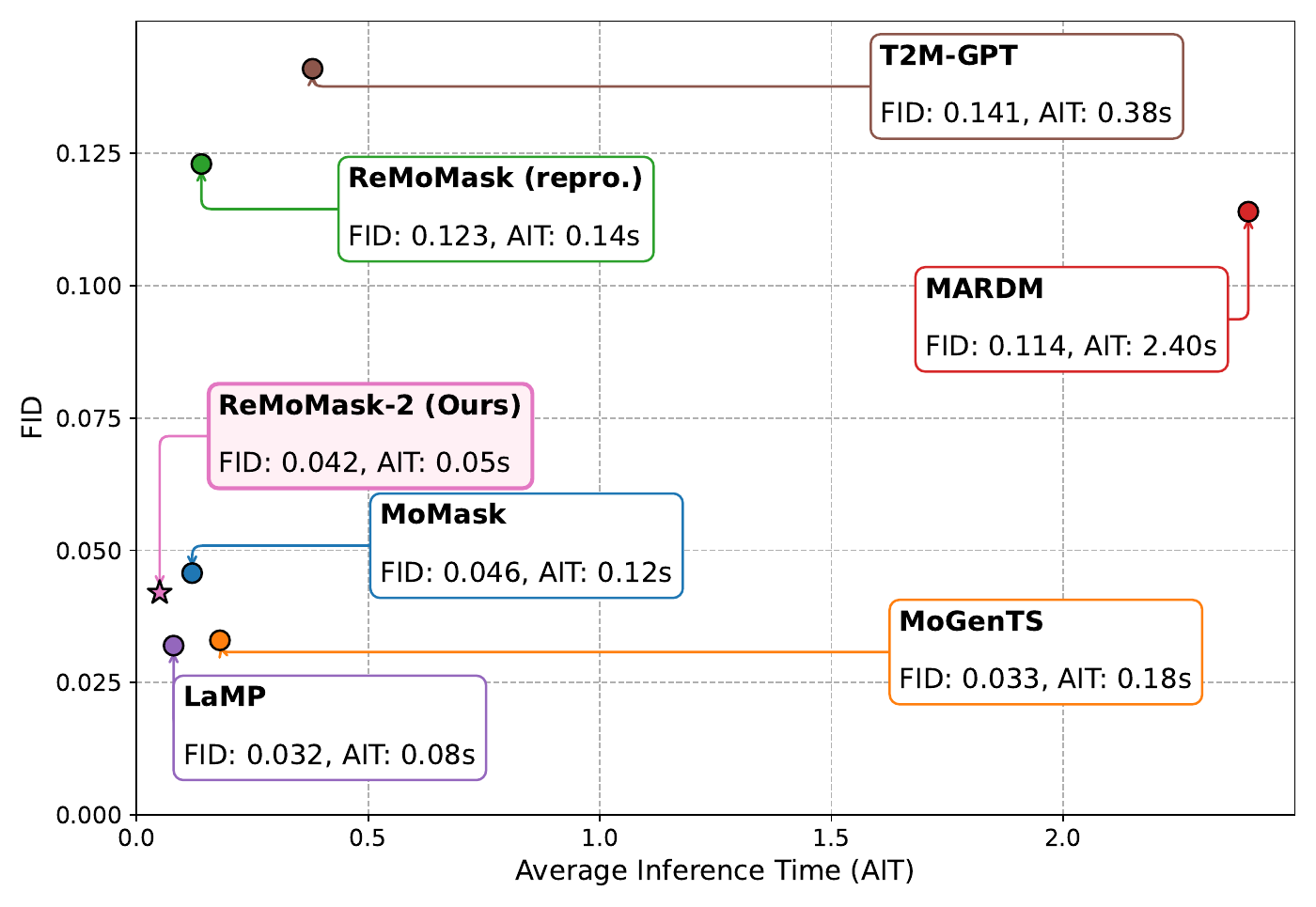}
    \vspace{-10pt}
    \caption{Comparison on FID and Inference Cost.}
    \vspace{-10pt}
    \label{fig:scatter}
\end{figure}

\noindent \myparagraph{Evaluation of Motion Generation.}
We evaluate motion generation performance on the HumanML3D, KIT-ML, and SnapMoGen datasets, as illustrated in Table~\ref{tab:t2m_experiment}. Evaluated under the unified protocol of Sec.~\ref{subsec:impl}, ReMoMask-2 is the strongest retrieval-augmented system on all three benchmarks against both published and reproduced baselines, attaining both the best FID and the best R-precision within that group, and it generally improves substantially over our conference-version ReMoMask on fidelity, R-precision, and MM-Dist across the three benchmarks, while using only a single mask-transformer stage. On HumanML3D it surpasses the full mask-plus-residual MoMask pipeline on both fidelity (0.042 versus 0.046) and Top-1 R-precision (0.528 versus 0.521) while itself running a single generative stage, and its fidelity is comparable to the strongest systems on this benchmark; on KIT-ML (0.138) and SnapMoGen (13.174) ReMoMask-2 attains the lowest FID among all compared methods. Multimodality decreases on HumanML3D and SnapMoGen while increasing on KIT-ML. We attribute the decrease on the first two benchmarks to retrieval conditioning narrowing the distribution of samples drawn for a given prompt, a hypothesis consistent with the trend on HumanML3D and SnapMoGen; on KIT-ML, where multimodality instead increases, a different mechanism is likely at play. It also delivers the fastest inference among the systems compared in Fig.~\ref{fig:scatter}.

\noindent \myparagraph{Qualitative Comparisons.}
We compare our method with others in Fig.~\ref{fig:rag_t2m_visual}. The qualitative examples illustrate that our model's motions align closely with the text descriptions, while the compared methods more often exhibit degraded motion quality or semantic mismatches; Sec.~\ref{sec:userstudy} reports a controlled user study quantifying this comparison.

\noindent\textbf{Inference Efficiency Comparison.}
Beyond generation quality, we compare our method with existing approaches in terms of both FID and inference cost. Following~\cite{momask}, the inference cost is quantified as the mean inference time over 100 samples on a single NVIDIA 2080Ti device, measured end to end and including, for the retrieval-augmented systems, the cost of encoding the query and searching the database. As shown in Fig.~\ref{fig:scatter}, ReMoMask-2 achieves the fastest inference among the retrieval-augmented and non-retrieval methods plotted there. Two changes contribute to this margin, only one of which is the removal of the residual-refinement stage: latent-aligned retrieval also replaces the conference-version retrieval stack (a full cross-modal encoder run over the prompt, followed by a search in a separate contrastive index) with a $1.57$M-parameter projector and a single cosine lookup over cached keys. Both changes act jointly to produce this speedup.

\subsection{Ablation Study}
\begin{figure}[t]
    \centering
    \includegraphics[width=\linewidth]{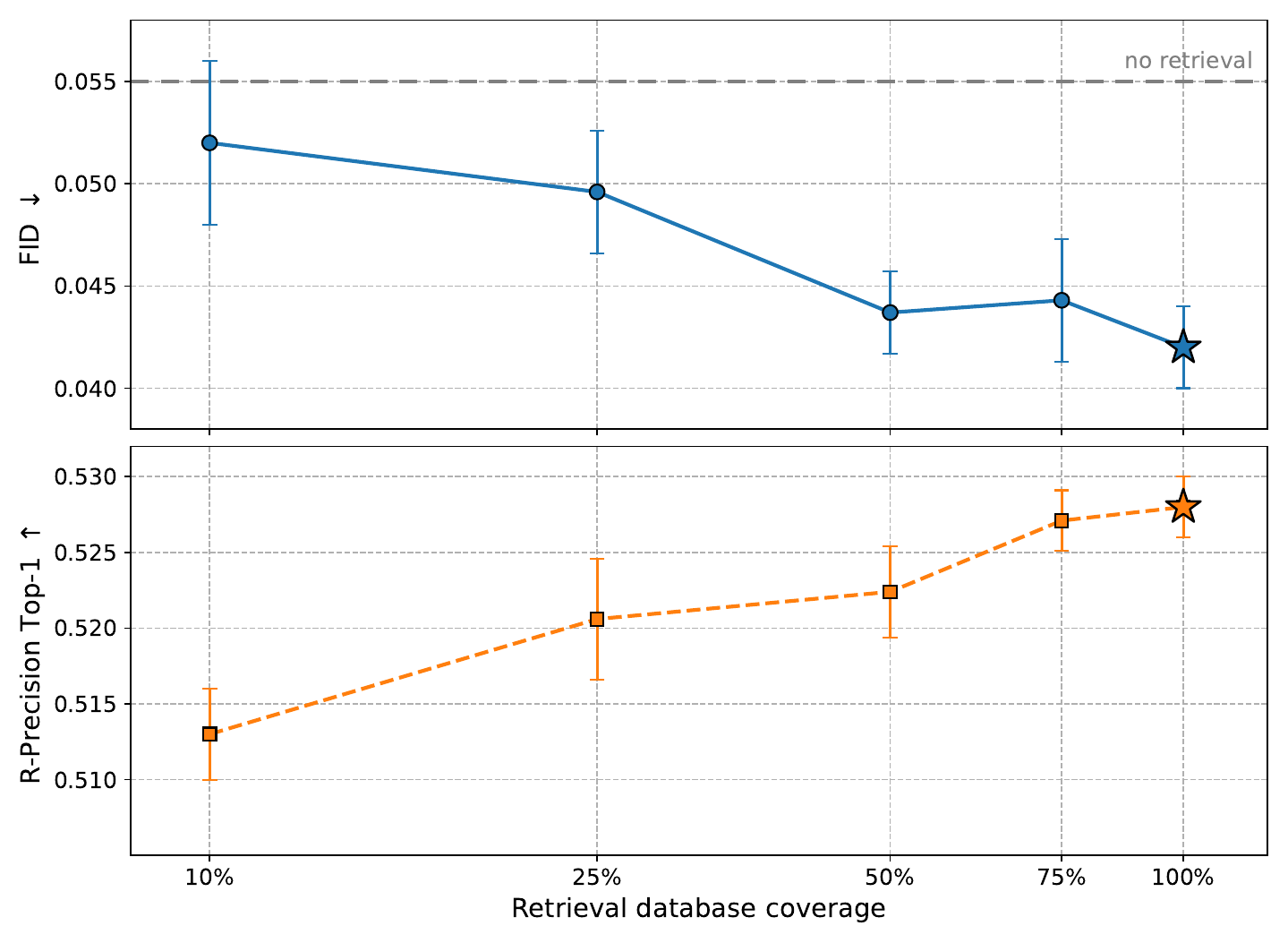}
    \vspace{-10pt}
    \caption{Effect of retrieval database coverage on HumanML3D. Subsets of the latent-aligned database are drawn uniformly at random with a shared seed across points and are swapped in at inference time only: the generator, the query projector, and the training recipe are those of the deployed model throughout, so these points share the intervention protocol of Table~\ref{tab:abl_ladder}. The dashed line marks the no-retrieval setting of that table; $100\%$ is the deployed configuration.}
    \vspace{-10pt}
    \label{fig:db_coverage_v2}
\end{figure}

Unless otherwise stated, every variant is trained and evaluated under the unified protocol of Section~\ref{subsec:impl} with single-stage mask-only inference; generation-quality numbers are therefore obtained under the same protocol as Table~\ref{tab:t2m_experiment}. Two questions inherited from the conference version are re-examined here, because latent alignment changes their premise: how much the conference components still contribute (Table~\ref{tab:abl_comp}), and how the benefit of retrieval scales with the size of the database (Fig.~\ref{fig:db_coverage_v2}). The remaining component-level ablations of the conference architecture (the hierarchical and momentum designs of the retriever, the full sweep over the information sources routed into the SSTA key and value pathways, the sensitivity of contrastive retriever training to its hyperparameters, and the cross-backbone transferability of the retriever) are reported in the conference version~\cite{remomask} and are not repeated here.

\begin{table}[t]
    \centering
    \caption{Impact of the retrieval representation on HumanML3D. The generator, fusion modules, and training recipe are identical across rows; only the representation in which evidence is retrieved and expressed changes. The second row keeps the conference-version database in the semantic space $\mathcal{S}$ but replaces the textual stream entering the fusion pathways with the latent-aligned $R_t = \phi(\cdot)$, so that the motion and textual streams can be moved into the generative geometry independently.}
    \label{tab:abl_space}
    \setlength{\tabcolsep}{8pt}
    \resizebox{0.95\columnwidth}{!}{
    \begin{tabular}{l|c c}
        \toprule
        Retrieval representation & FID$\downarrow$ & Top1$\uparrow$ \\
        \hline\hline
        Semantic space (HBM, conference design) & $0.109^{\pm.003}$ & $0.503^{\pm.003}$ \\
        Semantic space, latent-aligned $R_t$ & $0.101^{\pm.004}$ & $0.507^{\pm.003}$ \\
        Quantized indices ($z_q$) & $0.060^{\pm.003}$ & $0.519^{\pm.003}$ \\
        \rowcolor[gray]{0.90} Pre-quantization latent ($z_e$, Ours) & $\mathbf{0.042^{\pm.002}}$ & $\mathbf{0.528^{\pm.002}}$ \\
        \bottomrule
    \end{tabular}
    }
\end{table}

\noindent \myparagraph{Retrieval Space.}
Table~\ref{tab:abl_space} isolates the central design choice of ReMoMask-2. Keeping the single-stage generator and the SSTA fusion fixed, retrieving in the HBM semantic space (the conference retrieval design, retrained under the unified protocol) yields an FID of 0.109 for this generator paired with the conference retrieval space, whereas moving retrieval into the generator's own pre-quantization latent space improves it to 0.042, with Top1 following the same ordering. The two spaces indeed rank candidates very differently: over the test captions, each ranked against the full database of Section~\ref{subsec:lar}, the mean Spearman correlation between semantic-space and $z_e$-space rankings is 0.09 and their top-10 candidate sets overlap by 7.3\%, quantifying the retrieval--generation representation gap. Retrieving over quantized indices ($z_q$) recovers most but not all of the benefit (0.060), likely because discretization discards within-code variation that the continuous $z_e$ keys preserve.

This gain reflects an interaction between the two retrieval streams: their combined effect exceeds the sum of their individual, separable improvements. The second row of this table and the unaligned row of Table~\ref{tab:abl_route} each move one stream at a time, and the four configurations complete a $2{\times}2$: with both streams in the semantic space FID is 0.109; expressing only the retrieved caption in the generative geometry gives 0.101; rebuilding only the motion database there gives 0.098; expressing both there gives 0.042. Each move alone yields a modest reduction: expressing only the caption in the generator's geometry lowers FID from 0.109 to 0.101, and rebuilding only the motion database there lowers it to 0.098. But moving the caption after the database has already moved lowers FID by 0.056, from 0.098 to 0.042, far more than the 0.008 it is worth on its own: the two streams interact synergistically, and the full gain appears only once both are expressed together. This is what representation consistency means operationally: the evidence must arrive jointly, as a whole, in the generator's geometry.

\begin{table}[t]
    \centering
    \caption{Contribution of the conference-version components under latent-aligned retrieval, on HumanML3D. Each row is a separately trained variant: TSM is replaced by a uniform random masking schedule, and SSTA by a plain cross-attention fusion block of matched capacity that attends over $R_m$ and $R_t$ as an unstructured token pair. Both retrieval streams remain available to that block; only the asymmetric Q-K-V routing over the generator's token grid is removed; and the retrieval space, the database, and the training recipe are left unchanged. Absolute values here are comparable only within this table; Table~\ref{tab:abl_ladder} instead applies inference-time interventions to a single deployed model.}
    \label{tab:abl_comp}
    \setlength{\tabcolsep}{8pt}
    \resizebox{0.85\columnwidth}{!}{
    \begin{tabular}{l|c c}
        \toprule
        Variant & FID$\downarrow$ & Top1$\uparrow$ \\
        \hline\hline
        \rowcolor[gray]{0.90} ReMoMask-2 (full) & $\mathbf{0.042^{\pm.002}}$ & $\mathbf{0.528^{\pm.002}}$ \\
        w/o TSM (uniform masking) & $0.058^{\pm.003}$ & $0.522^{\pm.003}$ \\
        w/o SSTA (plain cross-attention) & $0.067^{\pm.004}$ & $0.514^{\pm.003}$ \\
        \bottomrule
    \end{tabular}
    }
\end{table}

\noindent \myparagraph{Conference Components.}
Table~\ref{tab:abl_comp} asks whether the two components of the conference architecture remain load-bearing once retrieval is moved into the generator's latent space. They do, and they fail in different ways. Replacing the topology structured masking schedule with uniform random masking costs 0.016 FID (0.042 to 0.058); Top1 degrades as well, by 0.006 and beyond the intervals covering the two entries, but by less than half of what the fusion ablation costs. TSM shapes only the training-time masking distribution, so the ablated model still receives the full retrieved evidence at inference and affects fidelity more than alignment. Replacing SSTA with a plain cross-attention block of matched capacity is more damaging along both axes (0.067 FID, Top1 0.514). The retrieval keys are pooled vectors and carry no grid of their own; the plain block appends that evidence as extra tokens instead of anchoring the fusion in the generator's joint--time token map (Sec.~\ref{subsec:ssta}), so fidelity and alignment degrade together. The relative ordering of the two arms matches the cumulative ablation of the conference architecture~\cite{remomask}, in which SSTA also accounted for more of the gain than TSM. Both arms nevertheless cost less than the retrieval-space and routing choices of Tables~\ref{tab:abl_space} and~\ref{tab:abl_route}, which places the components as necessary support for the new retrieval design: the extension augments the conference architecture. The SSTA arm and the first row of Table~\ref{tab:abl_route} isolate different components (the former keeps both retrieval streams and removes only the routing structure, whereas the latter keeps the routing structure but withholds $R_t$ entirely), so the larger cost of withholding $R_t$ there is consistent with this fusion ablation. Because HBM is no longer the deployed retriever, its contribution to ReMoMask-2 is measured in its new role, as a distillation teacher, in Table~\ref{tab:abl_distill}.

\begin{table}[t]
    \centering
    \caption{Graded conditioning ablation on HumanML3D. The first three rows keep real retrieved motion latents and destroy only their relevance to the query; the next two replace the evidence with degenerate constants; the row below the rule removes the retrieval pathway entirely and lies outside this ladder. All rows are inference-time interventions on the deployed model.}
    \label{tab:abl_ladder}
    \setlength{\tabcolsep}{6pt}
    \resizebox{\columnwidth}{!}{
    \begin{tabular}{l|c c c}
        \toprule
        Condition & FID$\downarrow$ & Top1$\uparrow$ & MMDIST$\downarrow$ \\
        \hline\hline
        \rowcolor[gray]{0.90} Real retrieval (Ours) & $\mathbf{0.042^{\pm.002}}$ & $\mathbf{0.528^{\pm.002}}$ & $\mathbf{2.860^{\pm.009}}$ \\
        Shuffled pairing & $0.048^{\pm.002}$ & $0.521^{\pm.003}$ & $2.889^{\pm.006}$ \\
        Random neighbors & $0.056^{\pm.003}$ & $0.517^{\pm.003}$ & $2.898^{\pm.006}$ \\
        Zeroed features & $0.063^{\pm.004}$ & $0.521^{\pm.003}$ & $2.884^{\pm.009}$ \\
        Constant (mean) prior & $0.070^{\pm.005}$ & $0.516^{\pm.003}$ & $2.902^{\pm.006}$ \\
        \midrule
        No retrieval & $0.055^{\pm.005}$ & $0.509^{\pm.002}$ & $2.968^{\pm.007}$ \\
        \bottomrule
    \end{tabular}
    }
\end{table}

\noindent \myparagraph{Conditioning Fidelity.}
Table~\ref{tab:abl_ladder} tests whether the generator consumes the \emph{semantic content} of retrieved evidence or merely benefits from its presence. The decisive evidence is the semantic ladder formed by the first three rows, where the retrieved entries remain real motion latents drawn from the same database and only their relevance to the query is destroyed: quality degrades monotonically from real retrieval (0.042) through shuffled pairing (0.048) to random neighbors (0.056), with each adjacent FID pair separated by non-overlapping confidence intervals; Top1 and MM-Dist reproduce the same ordering, though the gap between the two degraded conditions lies within their intervals. This ordering is the signature of genuine semantic consumption: were retrieval acting as a mere statistical regularizer, degrading its content while preserving its distribution would leave quality unchanged. The claim rests on FID, which is the only column in which the ladder separates step by step; we read the other two as corroborating its direction rather than as independent evidence. Removing retrieval altogether lands at 0.055, statistically indistinguishable from random neighbors. On fidelity the two degenerate conditions are worse still: the dataset-mean prior falls behind even the no-retrieval setting by a margin its confidence interval does not cover (0.070 versus 0.055), while the zero vector (0.063) is at best no better than removing retrieval. That ordering does not carry over to the other two columns: on Top1 and MM-Dist the zero-vector condition is indistinguishable from the rows that receive degraded but real evidence, so the degenerate constants are a statement about fidelity, not about alignment.

\noindent \myparagraph{Database Coverage.}
Fig.~\ref{fig:db_coverage_v2} varies the second quantity retrieval depends on. The ladder above destroys the \emph{relevance} of the retrieved evidence at a fixed database size; here relevance is left intact and the database is thinned instead, which caps the relevance that is available to be retrieved at all. Quality improves with coverage on both metrics: FID moves from 0.052 at $10\%$ coverage to 0.042 at full coverage and Top1 from 0.513 to 0.528, with the two ends of each sweep separated by their confidence intervals. A straight line in $\log_{10}$ of coverage describes the swept range well---$R^2 = 0.93$ on FID, at a slope of about $0.010$ FID per decade of database size. This fit describes the measured range only; extending it into a scaling-law claim would require sweeping database sizes beyond the single decade tested here. Because no adjacent pair of points is separated by its confidence interval, and points beyond half coverage are mutually indistinguishable, the figure supports the end-to-end trend and the flattening of returns as a range-level pattern. The two experiments are complementary rather than redundant: the ladder isolates relevance, this sweep isolates supply, and both point away from a distribution-level regularizer account. At $10\%$ coverage the point estimate still outperforms the no-retrieval level of Table~\ref{tab:abl_ladder} ($0.052$ versus $0.055$).

\begin{table}[t]
    \centering
    \caption{Sensitivity to the number of retrieved entries $k$ on HumanML3D.}
    \label{tab:abl_k}
    \setlength{\tabcolsep}{8pt}
    \resizebox{0.9\columnwidth}{!}{
    \begin{tabular}{l|c c c c c}
        \toprule
        $k$ & 1 & \cellcolor[gray]{0.90}2 (Ours) & 3 & 4 & 6 \\
        \hline\hline
        FID$\downarrow$ & $\mathbf{0.037^{\pm.002}}$ & \cellcolor[gray]{0.90}$0.042^{\pm.002}$ & $0.044^{\pm.004}$ & $0.043^{\pm.003}$ & $0.045^{\pm.003}$ \\
        Top1$\uparrow$ & $0.520^{\pm.003}$ & \cellcolor[gray]{0.90}$\mathbf{0.528^{\pm.002}}$ & $0.526^{\pm.004}$ & $0.525^{\pm.003}$ & $0.527^{\pm.002}$ \\
        \bottomrule
    \end{tabular}
    }
\end{table}

\noindent \myparagraph{Retrieval Count.}
Table~\ref{tab:abl_k} sweeps the number of retrieved entries. The optimum is shallow: $k{=}1$ attains the lowest FID (0.037) at a consistent cost in text--motion alignment (Top1 drops from 0.528 to 0.520), while larger $k$ degrades FID only mildly and leaves alignment essentially flat. We adopt $k{=}2$ as the balanced default, trading 0.005 FID for 0.008 Top1 against $k{=}1$; the criterion is joint fidelity and alignment, applied to the same metric suite that Table~\ref{tab:t2m_experiment} reports, so the deployed 0.042 is deliberately not the lowest FID this model attains. The remaining inference hyperparameters (guidance scale, iteration count, sampling temperature) are inherited unchanged from the conference protocol.

\begin{table}[t]
    \centering
    \caption{Value-pathway routing under latent-aligned retrieval on HumanML3D. All rows are separately trained variants that share the same generator, database, and training recipe; retrieval runs in the latent key space throughout and only the content appended to the Value pathway changes. Row 2 appends a fixed random unit vector $\xi$ of the same width, a contentless control that isolates the effect of adding one more Value entry from the effect of what that entry carries. Row 3 appends the undistilled semantic-space embedding $R_t^{\mathcal{S}} \in \mathcal{S}$, width-matched by a trainable linear adapter and optimized jointly with the rest of the model.}
    \label{tab:abl_route}
    \setlength{\tabcolsep}{8pt}
    \resizebox{0.85\columnwidth}{!}{
    \begin{tabular}{l|c c}
        \toprule
        Value content & FID$\downarrow$ & Top1$\uparrow$ \\
        \hline\hline
        $\mathrm{concat}(z, R_m)$ & $0.081^{\pm.003}$ & $0.525^{\pm.003}$ \\
        $\mathrm{concat}(z, R_m, \xi)$ (random vector) & $0.086^{\pm.004}$ & $0.523^{\pm.003}$ \\
        $\mathrm{concat}(z, R_m, R_t^{\mathcal{S}})$ (unaligned) & $0.098^{\pm.004}$ & $0.512^{\pm.003}$ \\
        \rowcolor[gray]{0.90} $\mathrm{concat}(z, R_m, R_t)$ (Ours) & $\mathbf{0.042^{\pm.002}}$ & $\mathbf{0.528^{\pm.002}}$ \\
        \bottomrule
    \end{tabular}
    }
\end{table}

\noindent \myparagraph{Value-Pathway Routing.}
Table~\ref{tab:abl_route} verifies the routing decision revisited in Section~\ref{subsec:lar}, and its four rows settle the question inside a single system. Routing the semantic-space $R_t$ into the Value pathway is harmful: at 0.098 FID it is worse than routing no textual evidence at all (0.081), and it is the only row in which text--motion alignment degrades appreciably as well (Top1 0.512, against 0.525 with no textual stream in the pathway). Routing the latent-aligned $R_t$ is beneficial: FID falls by nearly half, from 0.081 to 0.042, while alignment is essentially unchanged (Top1 0.525 versus 0.528). The generator, the fusion block, the database, and the training recipe are identical across the four rows, so the verdict on this design choice reverses once the textual evidence is expressed in the generator's own geometry by the distilled projector $\phi$, rather than by a separately trained linear adapter. The contentless control separates the two candidate explanations for so large a move from so small an intervention: appending a fixed random vector of the same width, one extra Value entry, no information, does not help and mildly hurts (0.086), so the deployed row's gain is attributable specifically to the content of $R_t$, as distinct from the mere capacity of an additional entry. The degraded rows are not all the same failure. Withholding $R_t$ leaves the Value pathway with evidence that is genuine and in-domain, so alignment is largely preserved and the cost is confined to fidelity, which we attribute to fusion layers that never learn to reconcile motion evidence with its textual counterpart. Supplying the unaligned $R_t^{\mathcal{S}}$ instead requires every synthesis step to reconcile two incompatible geometries, which corrupts the conditioning content itself and degrades fidelity and alignment together: the reading Table~\ref{tab:abl_ladder} already supports for its degenerate constants, where forcing an out-of-distribution condition through the fusion layers costs more than removing the condition. Since the adapter is trained end-to-end with the rest of the model, the network is free in principle to learn to ignore this input; that it remains harmful is consistent with incompatible geometry. The damage also stops short of retrieving in the wrong space altogether (0.109 in Table~\ref{tab:abl_space}): here the retrieved neighbors are the correct ones and only the textual branch is foreign to the substrate. Rows in this table, and the retrieval-space variants of Table~\ref{tab:abl_space}, are separately trained, unlike the inference-time interventions of Table~\ref{tab:abl_ladder}, so their absolute FID reflects that separately-trained protocol, distinct from the no-retrieval row there (0.055).

\begin{table}[t]
    \centering
    \caption{Distillation design for the query projector on HumanML3D. The three rows above the rule share the HBM retriever as teacher and vary the distillation objective; the row below it keeps the KL objective and swaps the teacher for TMR~\cite{tmr}. TC@1 denotes top-1 consistency between each projector and \emph{its own} teacher, measured over 4{,}384 held-out captions (one per test motion), and is therefore comparable in level only within the HBM-teacher rows; it is reported as a diagnostic and carries no preferred direction, which is why its column has no arrow. FID is obtained by plugging each projector into the frozen generation stack, with the generator, the database, and all inference settings left untouched.}
    \label{tab:abl_distill}
    \setlength{\tabcolsep}{6pt}
    \resizebox{0.9\columnwidth}{!}{
    \begin{tabular}{l|c c}
        \toprule
        Objective / teacher & TC@1 & FID$\downarrow$ \\
        \hline\hline
        \rowcolor[gray]{0.90} KL, HBM teacher (Ours) & $0.030$ & $\mathbf{0.042^{\pm.002}}$ \\
        InfoNCE, HBM teacher & $0.051$ & $0.064^{\pm.003}$ \\
        MSE regression, HBM teacher & $0.048$ & $0.372^{\pm.004}$ \\
        \midrule
        KL, TMR teacher & $0.044$ & $0.071^{\pm.003}$ \\
        \bottomrule
    \end{tabular}
    }
\end{table}

\noindent \myparagraph{Distillation Design.}
Table~\ref{tab:abl_distill} ablates how the query projector is aligned, and the outcome cautions against reading a retrieval proxy as a proxy for generation. Among the rows that share the HBM teacher, consistency with the teacher and downstream fidelity order the objectives differently: InfoNCE attains the highest top-1 agreement (0.051) and MSE regression is close behind (0.048), yet plugging those projectors into the frozen generation stack yields FID 0.064 and 0.372, respectively, whereas soft top-$\kappa$ distillation, the lowest on consistency, at 0.030, is the only objective that reaches 0.042. Top-1 agreement is a deliberately strict statistic here. The two spaces order candidates almost independently (Spearman 0.09, Table~\ref{tab:abl_space}), so the teacher's single highest-ranked entry frequently falls outside the set reachable as an argmax in $z_e$, and what distillation can transfer is the graded structure over the candidate set rather than its top element. What the deployed projector must do is retrieve relevant motions, and it does: over 4{,}384 held-out captions against the full database, $\phi$ recovers the ground-truth motion of a held-out caption at R@1 12.4, R@5 24.9 and R@10 34.6 directly in the latent key space, still enough to drive the best generation quality in this table. Regression is the extreme case: caption-to-motion correspondence is many-to-many, so a hard single target induces label noise, and reproducing individual teacher decisions can come at the cost of the geometry around them; InfoNCE improves on regression but still supervises one positive per caption. The size of the regression failure needs an explanation of its own, since 0.372 lies far outside the range that any inference-time perturbation of the deployed model spans in Table~\ref{tab:abl_ladder}. It is a collapse of the retrieved neighborhood rather than a ranking error: the MSE projector's queries concentrate into a small region of the key space, and across the held-out captions its retrieved sets cover only $6.2\%$ of the distinct database entries the deployed projector reaches. Nearly every prompt is then conditioned on the same handful of exemplars, which is far more destructive than the uniformly random neighbors of Table~\ref{tab:abl_ladder}: those at least vary from prompt to prompt and leave the conditioning signal merely uncorrelated with the text, rather than constant across the dataset. The collapse is partial rather than total: top-1 agreement with the teacher remains at 0.048. The last row varies the teacher rather than the objective, and the same conclusion holds along that second axis: distilling the identical KL objective from the external TMR retriever~\cite{tmr} reproduces that teacher's own top-1 decisions at 0.044, yet generates worse, at FID 0.071. What carries over into generation is therefore how well the teacher ranks in the first place, independent of how faithfully the student reproduces the teacher's own decisions. The database remains in the $z_e$ space in that row, so the retrieved evidence stays in-domain and only the induced neighborhoods change: TMR ranks text--motion pairs less accurately than HBM on this benchmark (Table~\ref{tab:rag_experiment}). Read together with Table~\ref{tab:abl_space}, a weaker ranker operating in the generator's latent space (0.071) is still far better than the strongest ranker operating in the semantic space (0.109), consistent with our central claim that representation consistency between retrieval and generation matters more than the quality of the ranker itself; within the latent space, replacing this weaker teacher with HBM further improves FID from 0.071 to 0.042. This is where HBM earns its place in ReMoMask-2: as the teacher whose graded ranking the projector inherits, distinct from the deployed retriever role reported in Table~\ref{tab:rag_experiment}, so that within this distillation pipeline, the teacher's retrieval accuracy (Table~\ref{tab:rag_experiment}) contributes to generation quality rather than remaining a disconnected number. We therefore read TC@1 as a diagnostic of alignment rather than as a model-selection criterion, and select both the distillation objective and the teacher by downstream generation quality.

\begin{table}[t]
    \centering
    \caption{Contribution of the residual-refinement stage across systems on HumanML3D. $\Delta$ is the FID change from adding the residual stage. The MoMask and conference-version ReMoMask $+$Residual entries are the same values as in Table~\ref{tab:t2m_experiment}.}
    \label{tab:abl_stage}
    \setlength{\tabcolsep}{5pt}
    \resizebox{\columnwidth}{!}{
    \begin{tabular}{l|c c c}
        \toprule
        System & Mask-only & + Residual & $\Delta$ \\
        \hline\hline
        MoMask & $0.085^{\pm.003}$ & $0.046^{\pm.002}$ & ${\color{mygreen}-0.039}$ \\
        ReMoMask & $0.143^{\pm.005}$ & $0.123^{\pm.003}$ & ${\color{mygreen}-0.020}$ \\
        \rowcolor[gray]{0.90} ReMoMask-2 (Ours) & $\mathbf{0.042^{\pm.002}}$ & $0.068^{\pm.003}$ & ${\color{red}+0.026}$ \\
        \bottomrule
    \end{tabular}
    }
\end{table}

\begin{table}[t]
    \centering
    \caption{Reconstruction quality of the frozen 2D RVQ-VAE as a function of the number of decoded quantization layers, on the HumanML3D test set. The tokenizer is the pretrained model of~\cite{mogents}; these numbers are re-measured by us rather than quoted, using the same feature extractor, the same reference statistics and the same set of test motions as the generation FIDs of Table~\ref{tab:t2m_experiment}, so that reconstruction and generation FID are on a common scale. Reconstruction is deterministic, so no repetition-based confidence intervals are reported.}
    \label{tab:abl_depth}
    \setlength{\tabcolsep}{8pt}
    \resizebox{0.85\columnwidth}{!}{
    \begin{tabular}{l|c c}
        \toprule
        Decoding & Recon.\ FID$\downarrow$ & MPJPE (mm)$\downarrow$ \\
        \hline\hline
        \rowcolor[gray]{0.90} Base layer only (used by ReMoMask-2) & $0.064$ & $36.8$ \\
        2 layers & $0.027$ & $28.4$ \\
        3 layers & $0.014$ & $23.1$ \\
        All 6 layers & $0.005$ & $16.1$ \\
        \bottomrule
    \end{tabular}
    }
\end{table}

\noindent \myparagraph{Single-stage sufficiency.}
Tables~\ref{tab:abl_stage} and~\ref{tab:abl_depth} together explain why ReMoMask-2 omits the residual-refinement stage. For prior two-stage systems the residual stage is load-bearing: it improves MoMask by 0.039 FID and the conference-version ReMoMask by 0.020. ReMoMask-2's single mask-transformer stage, decoding only the base quantization layer, already reaches 0.042, surpassing both the conference-version ReMoMask full pipeline (0.123) and MoMask's two-stage result (0.046). Re-attaching a residual-refinement transformer to that stage actively hurts: FID moves from 0.042 to 0.068, a degradation roughly an order of magnitude larger than the confidence intervals involved. Reconstruction FID falls from 0.064 at the base layer to 0.005 with all six layers (Table~\ref{tab:abl_depth}); the returns diminish with depth, as the first residual layer removes the largest share of the base-layer quantization error and each further layer a smaller one. Our generated motions therefore score better (0.042) than the base-layer reconstructions through which they are decoded (0.064). The two numbers measure different things: FID is a distance between distributions rather than a per-sample fidelity, so the systematic quantization bias shared by all base-layer reconstructions displaces the reconstructed distribution as a whole, whereas the generator, trained to match the data distribution in token space, absorbs part of that displacement. The depth curve reads the same way: adding a single residual layer cuts reconstruction FID by $58\%$ but MPJPE by only $23\%$, consistent with one residual code already removing most of the shared distributional offset while per-sample error can only be refined code by code. The reconstruction floor thus bounds per-sample accuracy specifically, leaving distributional quality unconstrained by it, and the depth curve marks per-sample-accuracy headroom that a deeper decoding path could still unlock---most of it already within reach at two layers---though, as Table~\ref{tab:abl_stage} shows, re-attaching a residual transformer forfeits that headroom instead of reaching it.

\subsection{Qualitative Results}
\label{sec:qualitative}
\begin{figure*}
        \centering
    \includegraphics[width=1\linewidth]{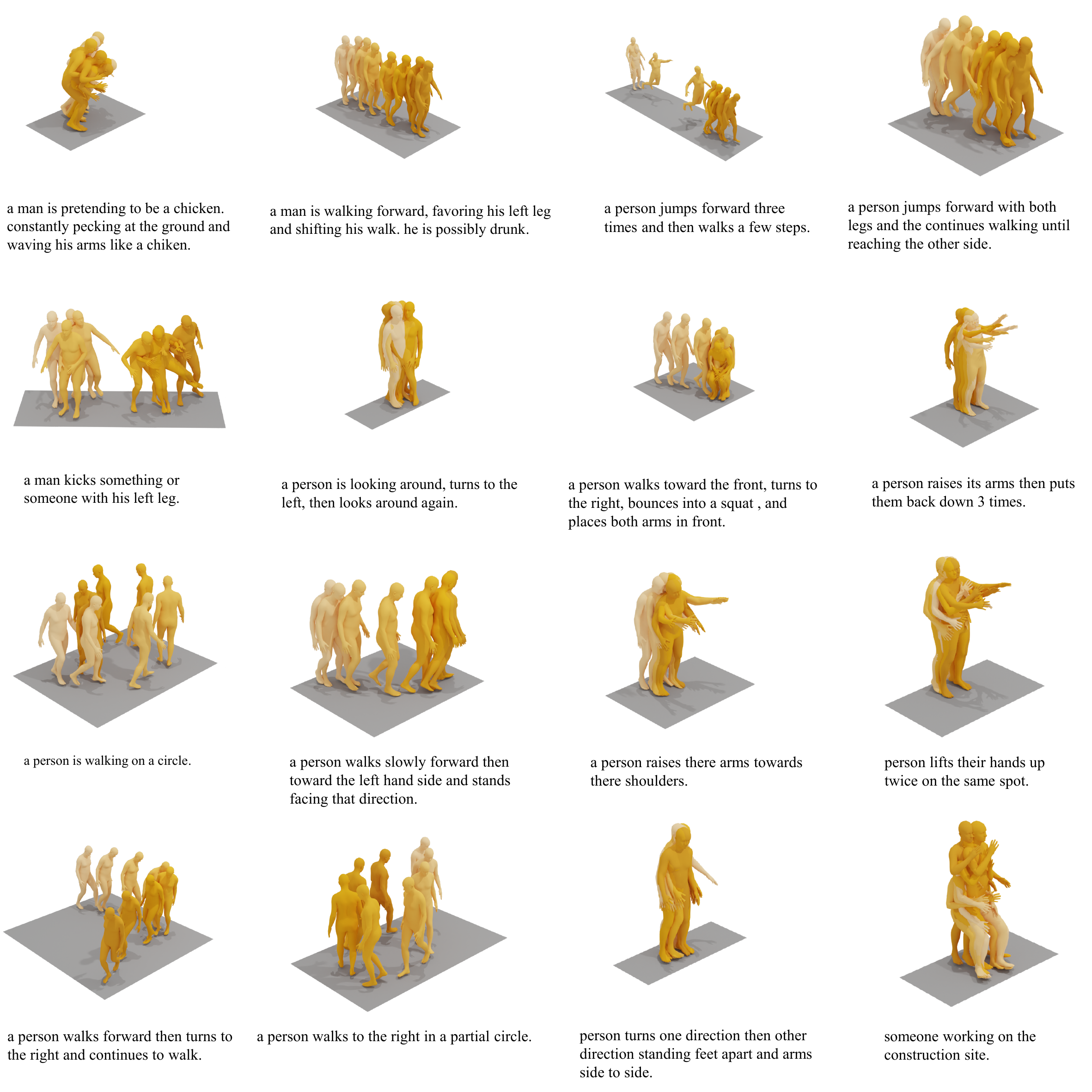}
    \caption{
We randomly sample and visualize 16 motions generated by the proposed ReMoMask-2 framework. These examples are conditioned on diverse prompts randomly selected from the HumanML3D~\cite{humanml3d}, providing qualitative evidence of the model’s ability to synthesize a wide range of realistic and semantically coherent motions.
    }
    \label{fig:demo}
\end{figure*}

\begin{figure*}
        \centering
    \includegraphics[width=1\linewidth]{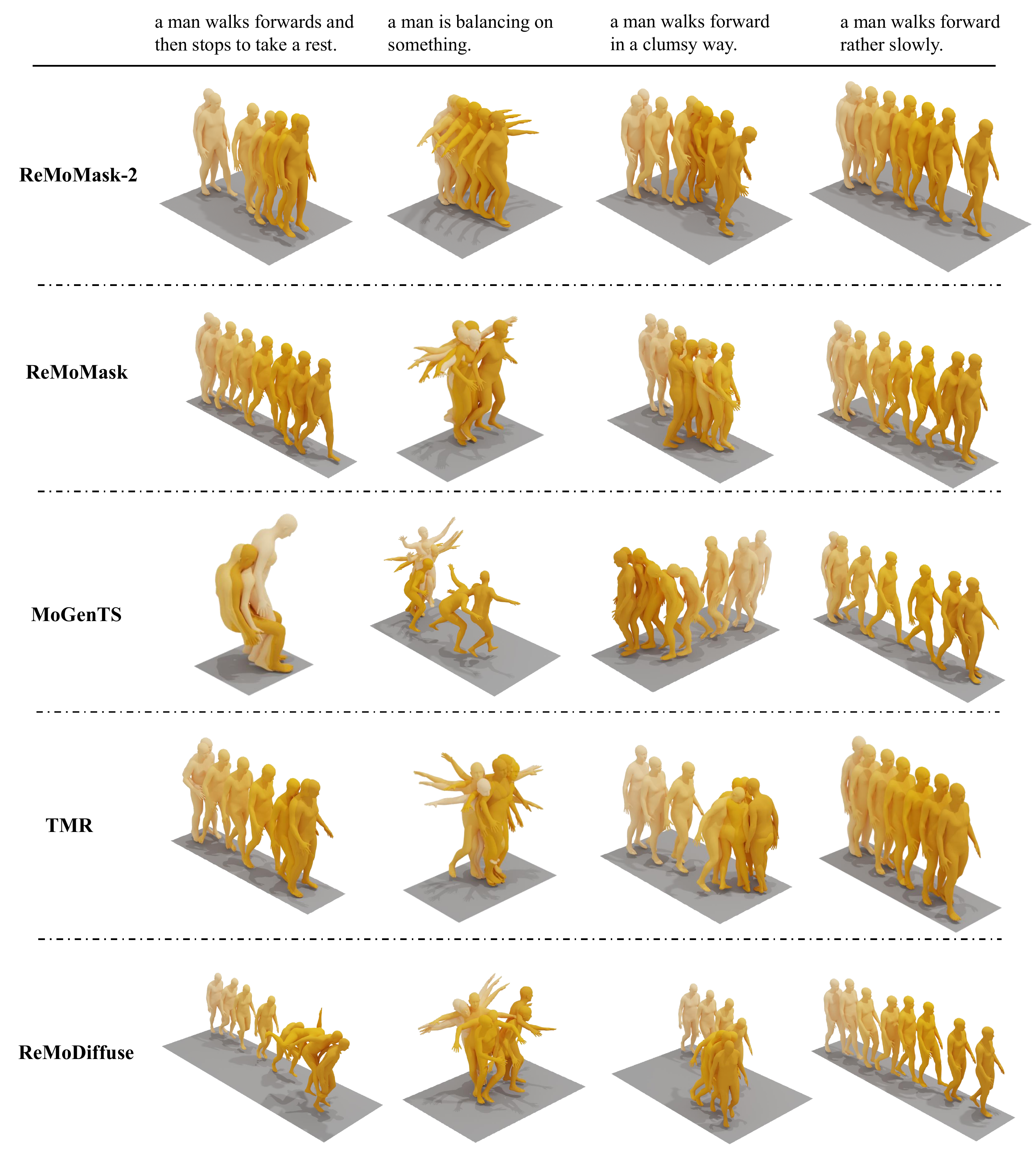}
    \caption{
    Comparison of the proposed ReMoMask-2 with ReMoMask and three state-of-the-art methods: MoGenTS~\cite{mogents}, TMR~\cite{tmr}, and ReMoDiffuse~\cite{remodiffuse}. We visualize motion sequences generated in response to four distinct text prompts. Each row represents the output of a different method, and each column corresponds to a specific prompt. The results demonstrate that ReMoMask-2 produces more realistic and semantically aligned motions compared to existing approaches.
    }
    \label{fig:demo1}
\end{figure*}

Fig.~\ref{fig:demo} illustrates ReMoMask-2's capability in generating diverse human motions. The 16 randomly inferred samples exhibit complex motion patterns such as directional transitions ("walks toward the front, turns to the right"), rhythmic actions ("raises arms three times"), and semantically rich behaviors ("pretending to be a chicken"), suggesting that the model captures nuanced motion dynamics and temporal transitions.
Fig.~\ref{fig:demo1} provides a comparative analysis of ReMoMask-2 against the conference-version ReMoMask, MoGenTS, TMR, and ReMoDiffuse. While baseline models generate basic motions like walking or balancing, our approach produces transitions that appear more natural (e.g., "walks forwards and then stops to take a rest" vs. simple linear motion) and physically plausible motion sequences (e.g., "walks forward in a clumsy way"); Sec.~\ref{sec:userstudy} reports a controlled user study quantifying motion quality and text-motion correspondence across methods.

\section{Limitations}
\label{sec:limitations}
The effectiveness of the proposed retrieval-augmented framework depends on the match between the motion database and the distribution a query is drawn from: when relevant motions are absent or sparsely represented in the retrieval corpus, the benefit that retrieval provides diminishes accordingly.
In addition, ReMoMask-2 decodes only the base quantization layer of the tokenizer, so its per-sample precision is bounded by base-layer reconstruction quality, leaving the finer detail carried by the deeper quantization layers untapped.
Extending latent-aligned retrieval to databases whose domain differs from the target one, detecting when a query falls outside the covered distribution, and unlocking the deeper decoding path within a single generative stage remain important directions for future work.

\section{Conclusion}
\label{sec:conclusion}

In this work, we present ReMoMask and its extension ReMoMask-2, retrieval-augmented masked generative frameworks for text-to-motion generation. To overcome the limitations of coarse-grained retrieval and ineffective fusion, we introduce \textbf{H}ierarchical \textbf{B}idirectional \textbf{M}omentum (HBM) for precise text-motion alignment and \textbf{T}opology \textbf{S}tructured \textbf{M}asking (TSM) to enforce structural consistency during training. Complemented by \textbf{S}emantic \textbf{S}patial-\textbf{T}emporal \textbf{A}ttention (SSTA) for knowledge integration, ReMoMask effectively bridges structured retrieval with high-quality generation. Building on this framework, ReMoMask-2 relocates retrieval into the generator's own pre-quantization latent space, closing the representation gap between retrieved evidence and the generative latents; a graded conditioning analysis confirms that the deployed model genuinely consumes the retrieved semantics rather than merely registering their presence. With retrieval acting in this shared space, a single mask-transformer stage, without a separate residual-refinement network, surpasses the accuracy of the full two-stage pipeline while reducing inference cost, yielding a system that is both more accurate and more efficient. Extensive experiments on HumanML3D, KIT-ML, and SnapMoGen confirm that our retriever delivers state-of-the-art text-to-motion retrieval, and that ReMoMask-2 consistently achieves the best generation fidelity among retrieval-augmented approaches, with the lowest FID among all compared methods on KIT-ML and SnapMoGen, while requiring only a single generative stage.

\noindent\textbf{Acknowledgements.}
This work was supported by the Fundamental Research Funds for the Central Universities, Peking University.

\appendix
\small
\section{Notations and Symbols}
\label{app:notation}

\begin{table}[t]
\centering
\caption{\textbf{Notations and symbols used in ReMoMask.}}
\setlength{\tabcolsep}{4pt}
\small
\renewcommand{\arraystretch}{1.1}
\resizebox{\linewidth}{!}{
\begin{tabular}{p{2.2cm} | l}
\toprule
\textbf{Symbol} & \textbf{Definition} \\
\midrule

$x$ & Input text prompt \\

$m$ & Motion sequence \\

$m^k$ & Motion of the $k$-th body part \\

$t_i$ & Text embedding of the $i$-th sample \\

$g_i$ & Global motion embedding \\

$p_{i,k}$ & Part-level motion embedding for the $k$-th body part \\

$R_m$ & Retrieved motion embedding \\

$R_t$ & Retrieved text embedding \\

$z$ & Latent motion tokens arranged on a $T \times J$ spatial--temporal grid \\

$z_{\mathrm{pred}}$ & Reconstructed latent motion tokens predicted by the generator \\

$h_{\mathrm{sem}}$ & Global semantic token constructed from text and retrieved context \\

$Q,K,V$ & Query, Key, and Value matrices in the attention module \\

$\alpha_{i,k}$ & Semantic relevance between text and the $k$-th body part \\

$\pi_{\mathrm{base}}$ & Base masking probability \\

$\pi_{i,k}$ & Masking probability of the $k$-th body part \\

$\mathcal{Q}$ & Momentum queue storing negative embeddings \\

$\mu$ & Momentum coefficient for updating momentum encoders \\

$\tau$ & Temperature parameter in contrastive learning \\

$\lambda_{P}$ & Weighting coefficient for the part-level contrastive loss \\

$B$ & Batch size \\

$K$ & Number of body parts \\

$T$ & Number of temporal frames \\

$J$ & Number of body joints \\

$N$ & Length of flattened motion tokens ($N=T \times J$) \\

$d$ & Dimension of the latent embedding space \\

$\mathrm{MLP}(\cdot)$ & Multilayer perceptron \\

$\mathrm{concat}(\cdot)$ & Concatenation operation \\

$\mathrm{flatten}(\cdot)$ & Flattening a 2D latent grid into a 1D sequence \\

\midrule

$\mathcal{S}, \mathcal{Z}$ & Contrastive semantic space; generative latent space \\

$\psi$ & Implicit cross-space translation $\mathcal{S} \rightarrow \mathcal{Z}$ in ReMoMask \\

$\mathcal{E}$ & Frozen 2D RVQ-VAE encoder producing the pre-quantization latent \\

$z_e$ & Pre-quantization continuous latent of the frozen RVQ-VAE encoder \\

$\bar{z}_e$ & Pooled, $\ell_2$-normalized $z_e$ used as a retrieval key \\

$d_e$ & Channel dimension of $z_e$ ($d_e = 1024$) \\

$\mathcal{D}$ & Latent-aligned retrieval database of $(\bar{z}_e, x)$ pairs \\

$\phi$ & Query projector mapping CLIP text embeddings into the $z_e$ space \\

$\kappa$ & Number of teacher candidates in distillation (top-$\kappa$) \\

$\Omega(x)$ & Top-$\kappa$ teacher candidate set for caption $x$ \\

$p^{\mathrm{HBM}}, p^{\phi}$ & Teacher / student retrieval distributions in KL distillation \\

$\mathcal{L}_{\mathrm{align}}$ & KL distillation objective for the query projector \\

$t$ & CLIP text embedding of a single prompt (single-sample form of $t_i$) \\

$T', J'$ & Temporal / spatial size of the downsampled $z_e$ grid ($T'{=}T/4,\, J'{=}6$) \\

$s^{\mathrm{HBM}}_j, s^{\phi}_j$ & Teacher / student similarity scores over candidate $j$ \\

\bottomrule
\end{tabular}
}
\label{tab:notation}
\end{table}

We list notations and symbols used in this paper, as shown in Table~\ref{tab:notation}.

\section{User Study}
\label{sec:userstudy}

To comprehensively evaluate the generation capability of \textbf{ReMoMask}, we conducted a comparative user study.  
We randomly selected 20 text prompts from the HumanML3D test set and generated motion sequences using \textbf{ReMoMask}, current state-of-the-art retrieval-augmented method (ReMoDiffuse), generative model (MoMask), and ground truth motions.

\begin{figure}[!ht]
\centering
    \includegraphics[width=1\linewidth]{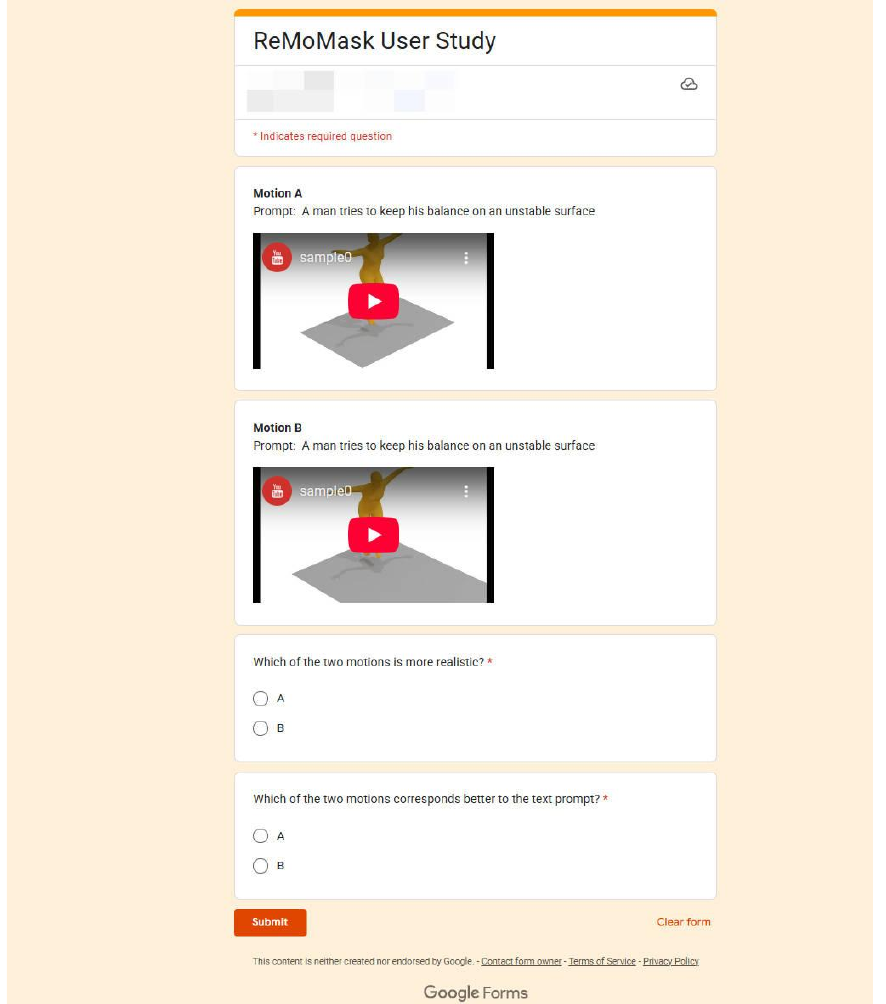}
    \caption{
    This figure illustrates the User Interface (UI) used in the ReMoMask User Study. Participants are presented with two motion videos, labeled as Motion A and Motion B, alongside a shared textual prompt. The motion clips are sampled from outputs generated by different models or the ground truth (GT), with model identities anonymized and video order randomized. Participants are asked to answer two evaluative questions: (1) “Which of the two motions is more realistic?”, assessing the visual plausibility and motion quality; and (2) “Which of the two motions corresponds better to the text prompt?”, evaluating the semantic alignment between the motion and the given description. This dual-question design enables a comprehensive human assessment of both motion realism and text-motion correspondence.
    }
    \label{fig:UI}
    \vspace{-2em}
\end{figure}

We employ a forced-choice paradigm in our user study, asking participants two key questions: “Which of the two motions is more realistic?” and “Which of the two motions corresponds better to the text prompt?”. The study is conducted via a Google Forms interface, as illustrated in Fig.~\ref{fig:UI}. To ensure fairness and reduce potential bias, the names of the generative models are hidden, and the order of presentation is randomized for each question. In total, over 50 participants took part in the evaluation.

\begin{figure}[!htb]
    \centering
    \includegraphics[width=\linewidth]{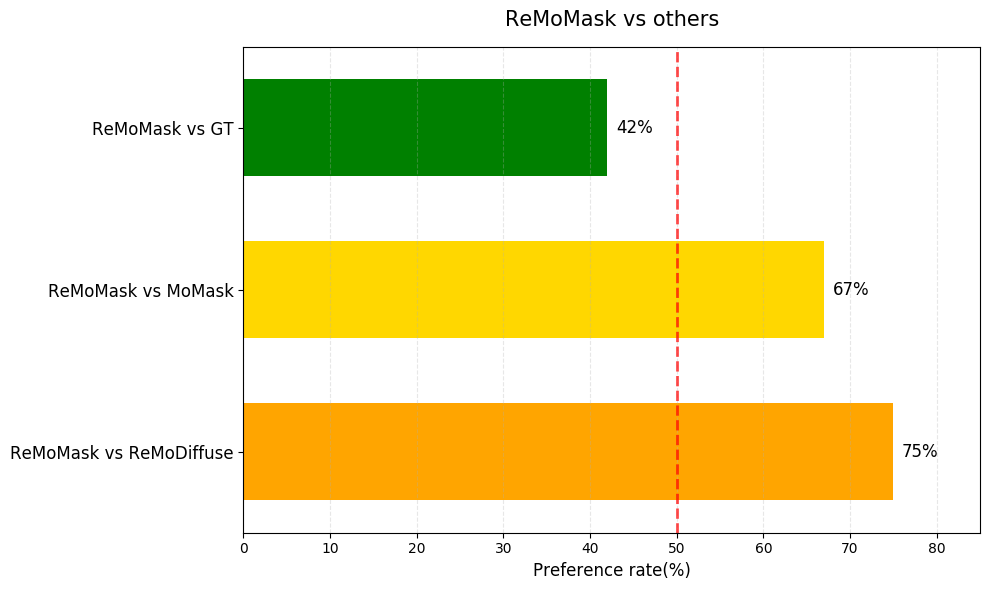}
    \caption{
      Motion Quality User Study
    }
    \label{fig:picture1}
\end{figure}

\begin{figure}[!htb]
    \centering
    \includegraphics[width=\linewidth]{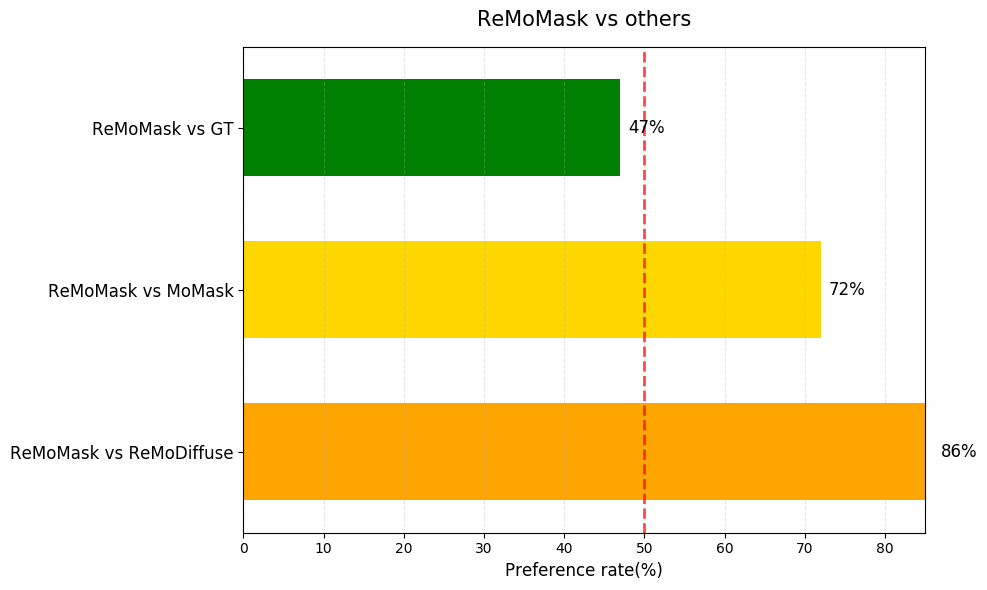}
    \caption{
    Text-Motion Correspondence User Study
    }
    \label{fig:picture2}
\end{figure}

Empirical results, depicted in Fig.~\ref{fig:picture1} and Fig.~\ref{fig:picture2}, underscore ReMoMask’s strong capability to generate motions that are not only realistic but also closely aligned with textual descriptions. Specifically, as shown in Fig.~\ref{fig:picture1}, ReMoMask achieves a 42\% preference rate over ground truth (GT) in terms of realism. Although GT motions are derived from real human data, this result indicates that ReMoMask is perceived as comparably realistic by human evaluators. Moreover, the model significantly outperforms both baselines: it achieves 67\% preference over MoMask and 75\% over ReMoDiffuse, demonstrating its strength in producing high-quality, lifelike motion sequences.

In terms of text correspondence (reported in Fig.~\ref{fig:picture2}), ReMoMask attains a 47\% preference rate over GT, suggesting that its generated motions exhibit nearly human-level alignment with text prompts. Compared to the baselines, ReMoMask again shows substantial improvements, with 72\% preference over MoMask and 86\% over ReMoDiffuse.

\begin{figure}[!htb]
    \centering
    \includegraphics[width=0.80\linewidth]{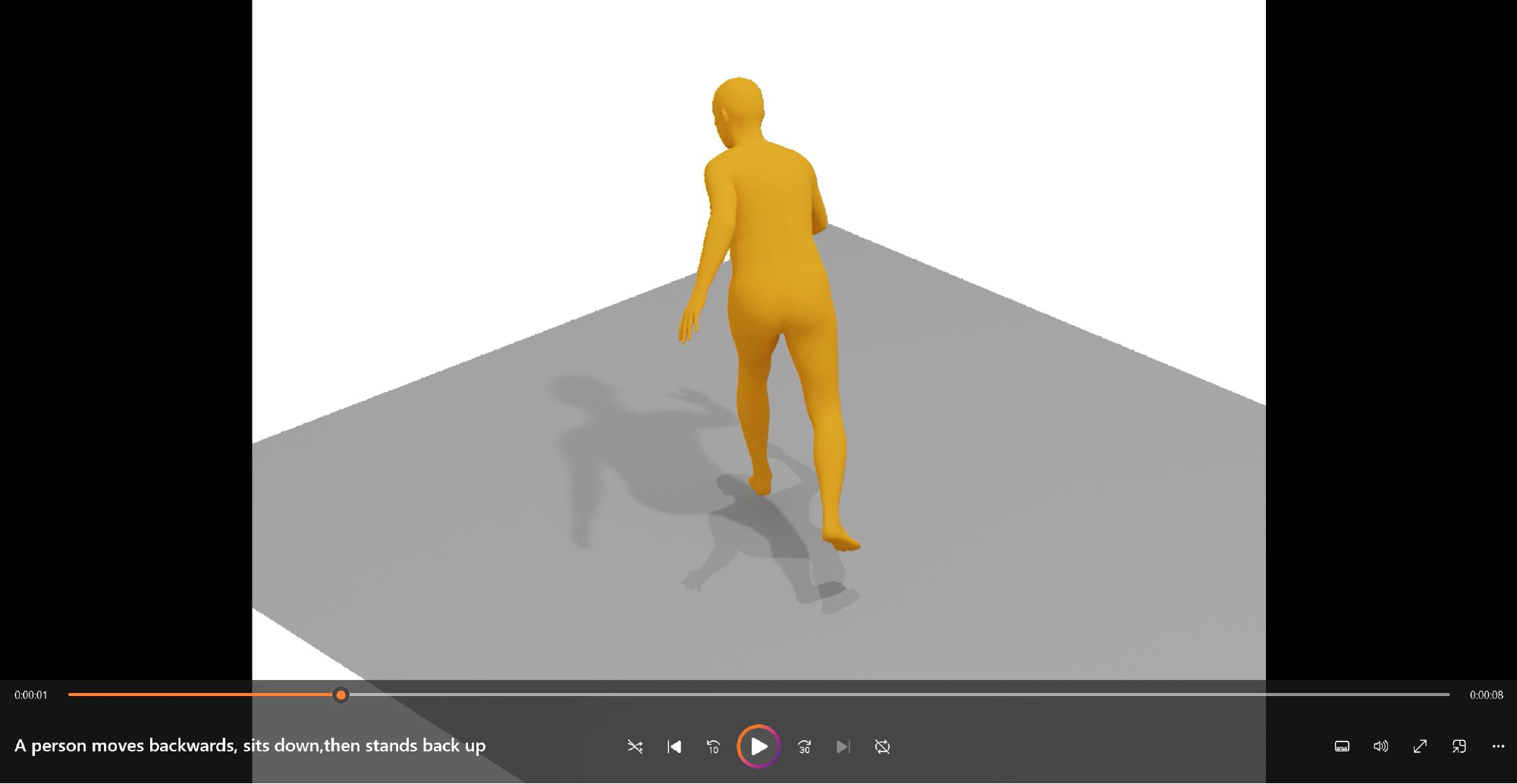}
    \vspace{-6pt}
    \caption{Video sample.}
    \vspace{-10pt}
    \label{fig:video_sample}
\end{figure}

\noindent\textbf{Video Demonstrations.}
We provide video demonstrations of our generated motions to facilitate qualitative evaluation. Fig.~\ref{fig:video_sample} shows representative video samples, where our method produces temporally coherent and semantically aligned motion sequences.

\bibliographystyle{IEEEtran}
\bibliography{reference}

\begin{IEEEbiography}[{\includegraphics[width=1in,height=1.25in,clip,keepaspectratio]{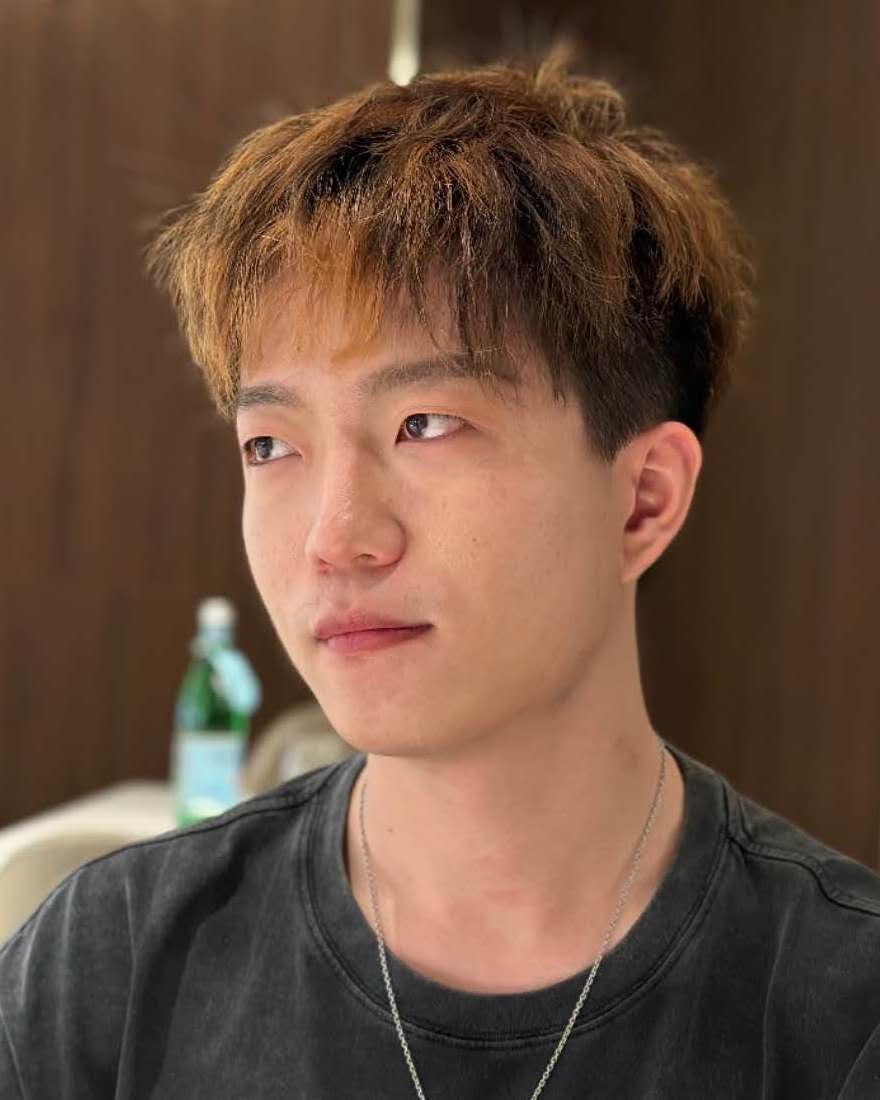}}]{Yiran Wang}
is a PhD student at the University of Sydney, Australia, advised by Dr Viorela Ila. He received his bachelor's and master's degrees from the University of Sydney. His research interests include computer vision and embodied AI, with a focus on human motion generation, retrieval-augmented generation, multimodal representation learning, and robotics.
\end{IEEEbiography}

\begin{IEEEbiography}[{\includegraphics[width=1in]{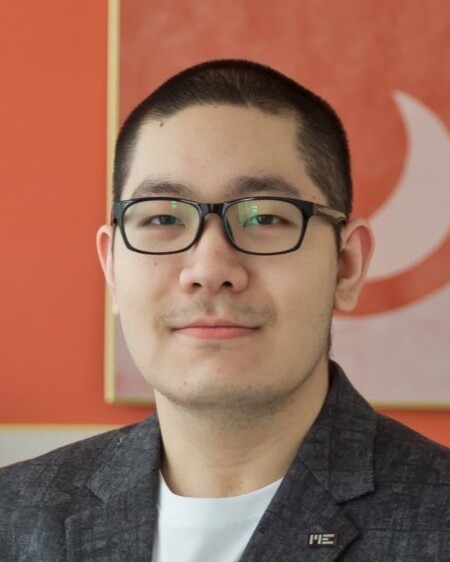}}]{Zeyu Zhang}
is a researcher working on generative AI, with a particular interest in building models that understand and interact with the physical world. He received his bachelor’s degree from the Australian National University, where he was advised by Prof. Richard Hartley and Prof. Ian Reid. His research explores generative modeling for learning physical dynamics from visual data. His work spans world models, multimodal foundation models, embodied AI, and AI for health. 
\end{IEEEbiography}

\begin{IEEEbiography}[{\includegraphics[width=1in,height=1.25in,clip,keepaspectratio]{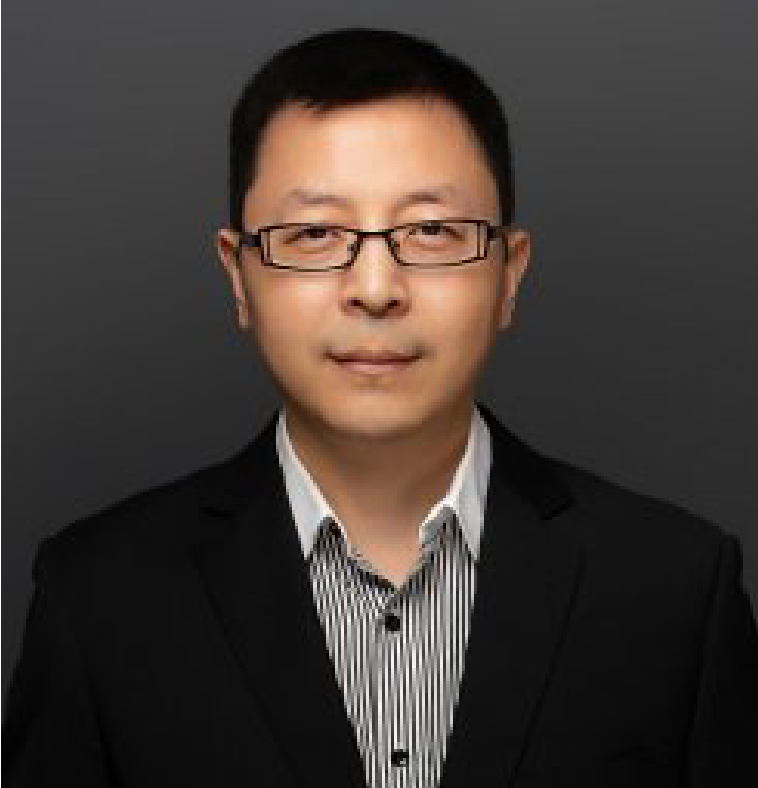}}]{Ling Shao}
(Fellow, IEEE) is a Distinguished Professor with the University of Chinese Academy of
Sciences, Beijing, China. He was the founder of the
Inception Institute of Artificial Intelligence (IIAI)
and the Mohamed bin Zayed University of Artificial
Intelligence (MBZUAI), Abu Dhabi, UAE. His research interests include physical AI, multimodal AI,
and AI for healthcare. He is a fellow of the IEEE,
the IAPR, the BCS and the IET.
\end{IEEEbiography}

\begin{IEEEbiography}[{\includegraphics[width=1in,height=1.25in,clip,keepaspectratio]{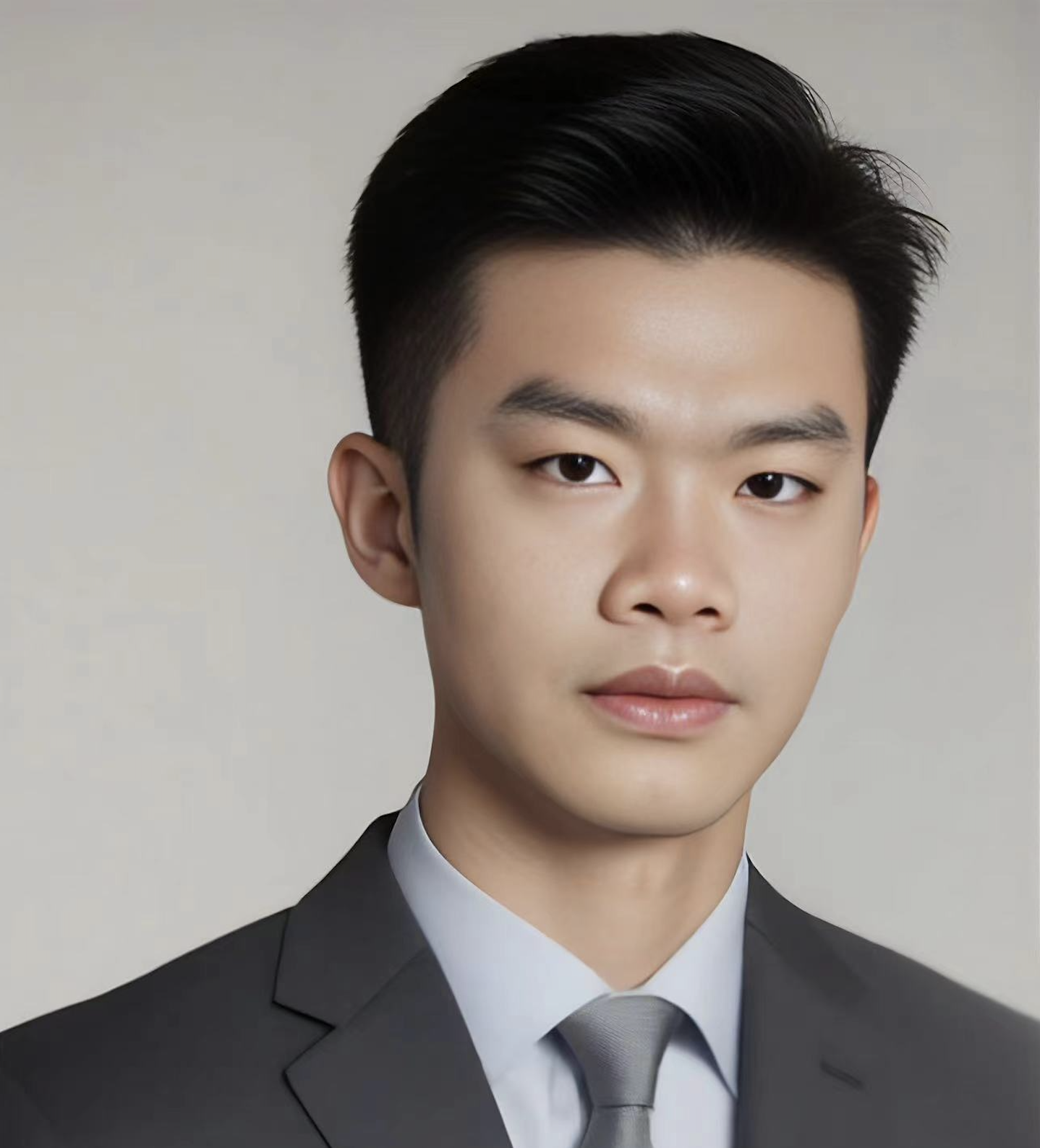}}]{Hao Tang}
is an Assistant Professor at Peking University, China. Previously, he held postdoctoral positions at CMU, USA, and ETH Zürich, Switzerland. He earned his master’s degree from Peking University, and his Ph.D. from the University of Trento, Italy.
He has had the opportunity to visit the University of Oxford, Northeastern University, NUS, and IIAI, among other institutions.
His research interests include computer vision, generative AI, spatial intelligence, world model, and embodied AI.
\end{IEEEbiography}

\end{document}